%% file: main.tex
\documentclass[a4paper]{scrartcl}
\typearea{13}
\usepackage{ae}            
\usepackage{times}          

\usepackage[utf8]{inputenc} 
\usepackage[T1]{fontenc}    
\usepackage{amsfonts}       
\usepackage{graphicx}
\usepackage{amsmath,amssymb} 
\usepackage{esint}  
\usepackage{xcolor}
\usepackage{dsfont}        

\newcommand{\R}{\mathds{R}}
\renewcommand{\vec}[1]{\boldsymbol{\mathbf{#1}}}
\providecommand{\abs}[1]{\lvert#1\rvert}
\providecommand{\norm}[1]{\lVert#1\rVert}
\usepackage{exscale}       
\usepackage{hyperref} 
\usepackage[all]{hypcap} 
\definecolor{darkred}{rgb}{0.5,0,0}
\definecolor{darkgreen}{rgb}{0,0.5,0}
\definecolor{darkblue}{rgb}{0,0,0.5}

\usepackage{tabularx} 

\hypersetup{linktocpage
,bookmarksnumbered=true
,colorlinks
,linkcolor=darkblue
,filecolor=darkgreen
,urlcolor=darkred
,citecolor=darkblue
,breaklinks=true
,pdfauthor={Stanislav Nikolov, Sam Blackwell, Ruheena Mendes, Michelle Livne, Yojan Patel, Alexei Zverovitch, Jeffrey De Fauw, Clemens Meyer, Harry Askham, Bernardino Romera-Paredes, Alan Karthikesalingam, Carlton Chu, Dawn Carnell, Cheng Boon, Derek D'Souza, Syed Ali Moinuddin, Bethany Garie, Yasmin McQuinlan, Sarah Ireland, Kiarna Hampton, Krystle Fuller, Hugh Montgomery, Geraint Rees, Mustafa Suleyman, Trevor Back, Joseph R. Ledsam, Cían Hughes}
,pdftitle={Deep learning to achieve clinically applicable segmentation of head and neck anatomy for radiotherapy}
,pdfsubject={Progress Report}
,raiselinks
}

\usepackage{scrlayer-scrpage}
\setheadsepline{0.4pt}
\ohead{\thepage}
\setcapindent{0em}
\setkomafont{captionlabel}{\footnotesize\sffamily\bfseries}
\setkomafont{caption}{\footnotesize\sffamily}
\setkomafont{author}{\sffamily}

\usepackage{longtable}
\usepackage{multirow}
\usepackage{booktabs}  
\usepackage{colortbl}
\usepackage{siunitx}

\usepackage{authblk} 
\usepackage{pdflscape}

\begin{document}

\title{Deep learning to achieve clinically applicable segmentation of head and neck anatomy for radiotherapy}
\author[1*]{Stanislav~Nikolov}
\author[1*]{Sam~Blackwell}
\author[2*]{Alexei~Zverovitch}
\author[3]{Ruheena~Mendes} 
\author[2]{Michelle~Livne}
\author[1]{Jeffrey~De~Fauw}
\author[2]{Yojan~Patel}
\author[1]{Clemens~Meyer}
\author[1]{Harry~Askham}
\author[1]{Bernardino~Romera-Paredes}
\author[2]{Christopher~Kelly}
\author[2]{Alan~Karthikesalingam}
\author[1]{Carlton~Chu}
\author[3]{Dawn~Carnell}
\author[4]{Cheng~Boon} 
\author[3]{Derek~D'Souza}
\author[3]{Syed~Ali~Moinuddin}
\author[1]{Bethany Garie}
\author[1]{Yasmin McQuinlan}
\author[1]{Sarah Ireland}
\author[1]{Kiarna Hampton}
\author[1]{Krystle Fuller}
\author[2,5,6]{Hugh~Montgomery}
\author[2,5]{Geraint~Rees}
\author[1]{Mustafa~Suleyman}
\author[1]{Trevor~Back}
\author[2,3+]{Cían~O.~Hughes}
\author[7+]{Joseph~R.~Ledsam}
\author[1+]{Olaf~Ronneberger}
\affil[1]{DeepMind, London, UK}
\affil[2]{Google Health, London, UK}
\affil[3]{University College London Hospitals NHS Foundation Trust, London, UK}
\affil[4]{Clatterbridge Cancer Centre NHS Foundation Trust, Liverpool, UK}
\affil[5]{University College London, London, UK}
\affil[6]{Centre for Human Health and Performance, and Institute for Sports, Exercise and Health, University College London, London, UK}
\affil[7]{Google AI, Tokyo, Japan}
\affil[*]{These authors contributed equally to this work}
\affil[+]{These authors contributed equally to this work}
\date{}
\maketitle

\begin{abstract}
Over half a million individuals are diagnosed with head and neck cancer each year worldwide.
Radiotherapy is an important curative treatment for this disease, but it requires manual time consuming delineation of radio-sensitive organs at risk (OARs).
This planning process can delay treatment, while also introducing inter-operator variability with resulting downstream radiation dose differences. While auto-segmentation algorithms offer a potentially time-saving solution, the challenges in defining, quantifying and achieving expert performance remain.
Adopting a deep learning approach, we demonstrate a 3D U-Net architecture that achieves expert-level performance in delineating 21 distinct head and neck OARs commonly segmented in clinical practice.
The model was trained on a dataset of 663 deidentified computed tomography (CT) scans acquired in routine clinical practice and with both segmentations taken from clinical practice and segmentations created by experienced radiographers as part of this research, all in accordance with consensus OAR definitions.
We demonstrate the model’s clinical applicability by assessing its performance on a test set of 21 CT scans from clinical practice, each with the 21 OARs segmented by two independent experts. We also introduce surface Dice similarity coefficient (surface DSC), a new metric for the comparison of organ delineation, to quantify deviation between OAR surface contours rather than volumes, better reflecting the clinical task of correcting errors in the automated organ segmentations. The model's generalisability is then demonstrated on two distinct open source datasets, reflecting different centres and countries to model training. With appropriate validation studies and regulatory approvals, this system could improve the efficiency, consistency, and safety of radiotherapy pathways.

\end{abstract}


\section{Introduction}

Each year, 550,000 people are diagnosed with cancer of the head and neck worldwide \cite{Jemal2011-rq}.
This incidence is rising \cite{Cancer_Research_UK2018-bb}, more than doubling in certain subgroups over the last 30 years \cite{National_Cancer_Intelligence_Network2012-ny,Oxford_Cancer_Intelligence_Unit2010-yb,Parkin2011-rj}.
Where available, most will be treated with radiotherapy which targets the tumour mass and areas at high risk of microscopic tumour spread. 
However, strategies are needed to mitigate the dose-dependent adverse effects which result from incidental irradiation of normal anatomical structures (‘organs at risk’, OARs) \cite{Jensen2007-im,Dirix2009-lb,Caudell2010-ze,Nutting2011-ni}.

The efficacy and safety of head and neck radiotherapy thus depends upon the accurate delineation of OARs and tumour, a process known as segmentation or contouring.
However, the fact that this process is predominantly done manually means that results may be both inconsistent and imperfectly accurate \cite{Nelms2012-dv}, leading to  large inter- and intra-practitioner variability even amongst experts, and thus variation in care quality \cite{Voet2011-kw}.

\begin{figure}[htbp]
    \centering
    \includegraphics[width=\textwidth]{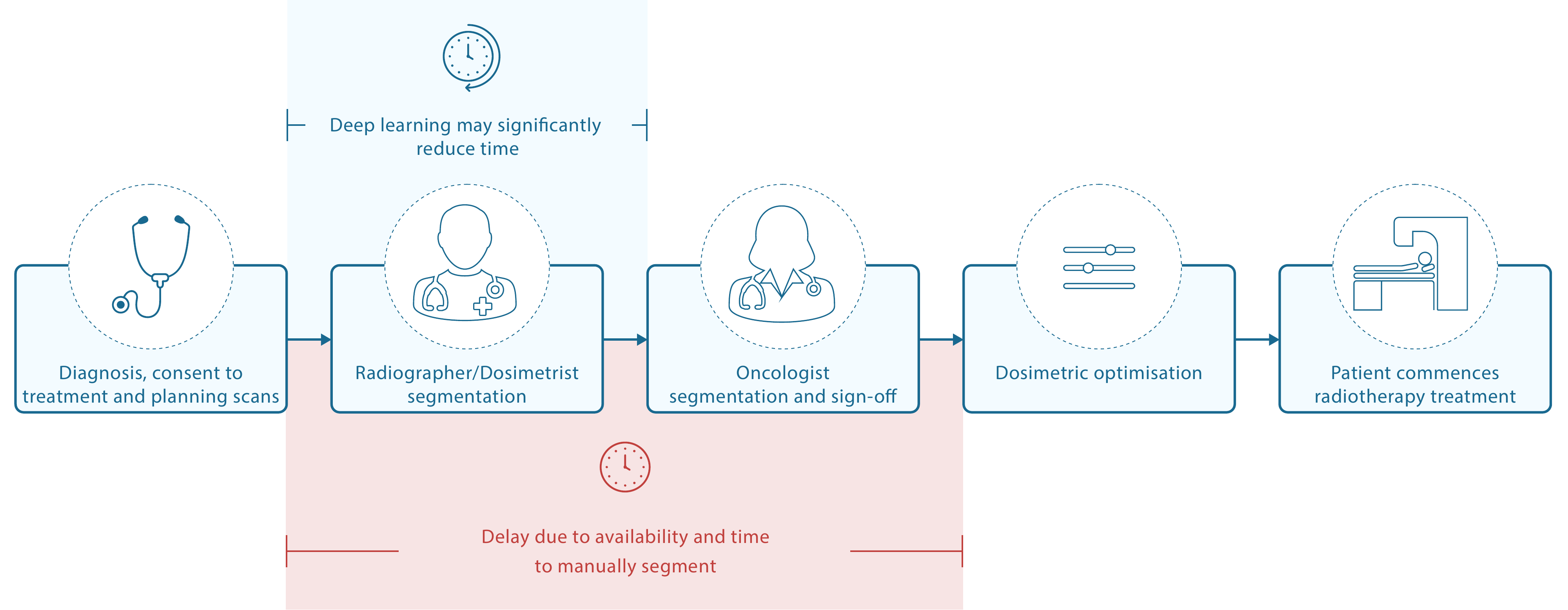}
    \caption{\textbf{
A typical clinical pathway for radiotherapy}.
After a patient is diagnosed and the decision is made to treat with radiotherapy, a defined workflow aims to provide treatment that is both safe and effective. In the UK the time delay between decision to treat and treatment delivery should be no greater than 31 days\cite{nhscancerplan}.
Time-intensive manual segmentation and dose optimisation steps can introduce delays to treatment.
    }
    \label{fig:pathway}
\end{figure}

Segmentation is also very time consuming: an expert can spend four hours or more on a single case \cite{Harari2010-sd}.
The duration of resulting delays to treatment commencement (see \autoref{fig:pathway}) is associated with increased risk both of local recurrence and of overall mortality \cite{Chen2008-fp,Mikeljevic2004-cc}. Increasing demands for, and shortages of, trained staff already place a heavy burden on healthcare systems which can lead to long delays for patients as radiotherapy is planned \cite{Round2013-xe,Rosenblatt_E2017-be}, and the continued rise in head and neck cancer incidence may make it impossible to maintain even current temporal reporting standards \cite{Oxford_Cancer_Intelligence_Unit2010-yb}. Such issues also represent a barrier to ’Adaptive Radiotherapy’- the process of repeated scanning, segmentation and radiotherapy planning throughout treatment which maintains the precision of tumour targeting (and OARs avoidance) in the face of treatment-related anatomic changes such as tumour shrinkage \cite{Veiga2014}.

Automated (i.e. computer-performed) segmentation has the potential to address these challenges. However, 
most segmentation algorithms in clinical use are atlas-based, producing their segmentations by fitting previously labelled reference images to the new target scan.
This might not sufficiently account for either post-surgical changes, or the variability in normal anatomical structure which exists between patients, particularly when considering the variable effect that tumours may have on local anatomy; they may thus be prone to systematic error. To date, such algorithm-derived segmentations still require significant manual editing, perform at expert levels on only a small number of organs, demonstrate an overall performance in clinical practice inferior to that of human experts, and have failed to significantly improve clinical workflows \cite{Daisne2013-uk,Fortunati2013-ay,Duc2015-qs,Hoogeman2008-dw,Levendag2008-ax,Qazi2011-ta,Sims2009-ou,Teguh2011-dq,Thomson2014-pr,Walker2014-dd,Gacha2018,Raudaschl2017-mn,WU2019}.

In recent years, deep learning based algorithms have proven capable of delivering substantially better performance than traditional segmentation algorithms. In head and neck cancer segmentation, several deep learning based approaches have been proposed. Some of them use standard convolutional neural network classifiers on patches with tailored pre- and post-processing \cite{Fritscher2016-nd,Ibragimov2017-is,Mocnik2018-fv,Ren2018-vi,zhong2019boosting}.
However, the U-Net convolutional architecture \cite{Ronneberger2015-ur} has shown promise in the area of deep-learning based medical image segmentation \cite{De_Fauw2018-tj} and has now also been applied to head and neck radiotherapy segmentation \cite{Hansch2018-jo,Zhu2018-js,Tong2018,Liang2018,willems2018-DeepVoxNet,kodym2018-ed,wang2019organ,men2019,tappeiner2019multi,rhee2019automatic,tang2019clinically,van2019deep,gou2020self,Mak-2020}.

Despite the promise deep learning offers, barriers remain to the application of auto-segmentation to radiotherapy planning. These include the absence of consensus on how 'expert' performance is defined, the lack of available methods by which such human performance can be compared to that delivered by automated segmentation processes, and thus how the clinical acceptability of automated processes can be defined.

Here we address these challenges, and report a deep learning approach that delineates a wide range of important OARs in head and neck cancer radiotherapy scans, to human expert standard.
We achieve this using a study design that includes 
(i) the introduction of a clinically meaningful performance metric for segmentation in radiotherapy planning; 
(ii) a representative set of images acquired during routine clinical practice; 
(iii) an unambiguous segmentation protocol for all organs; and 
(iv) a segmentation of each test set image according to these protocols by two independent experts. 
In addition to the model's generalisability, as demonstrated on two distinct open source datasets, by achieving performance equal to human experts on previously unseen patients from the same hospital site used for training we demonstrate the clinical applicability of our approach.

\section{Results}

\subsection{Selecting clinically representative datasets}
Datasets are described in detail in the Methods section. In brief, the first dataset was a representative sample of CT scans used to plan curative-intent radiotherapy of head and neck cancer for patients at University College London Hospitals NHS Foundation Trust (UCLH), a single high-volume centre. We performed iterative cycles of model development using the UCLH scans ('training' and 'validation' subsets), taking the performance on a previously unseen subset ('test') as our primary outcome.

It is also important to demonstrate a model's generalisability to data from previously unseen demographics and distributions. To do this we curated test and validation datasets of open source CT scans. These were collected from The Cancer Imaging Archive (“TCIA test set”) \cite{Bosch2015-tw, Clark2013-zx, Zuley2016-nb} and the "PDDCA": Public Domain Database for Computational Anatomy dataset released as part of the 2015 challenge (“PDDCA test set”; \cite{Raudaschl2017-mn}).

\autoref{tab:dataset_characteristics} details the characteristics of these datasets and their patient demographics. Ethnicity and protected-group status is not reported, as this information was not available in the source systems. Twenty-one organs at risk were selected to represent a wide range of anatomical regions throughout the head and neck. To provide a human clinical comparison for the algorithm, each case was manually segmented by a single radiographer with arbitration by a second radiographer. This was compared to our study's 'gold standard' ground truth graded by two further radiographers and arbitrated by one of two independent specialist oncologists, each with a minimum of four years specialist experience in radiotherapy treatment planning for head and neck patients. 

\begin{figure}[htbp]
    \centering
    \includegraphics[width=\textwidth]{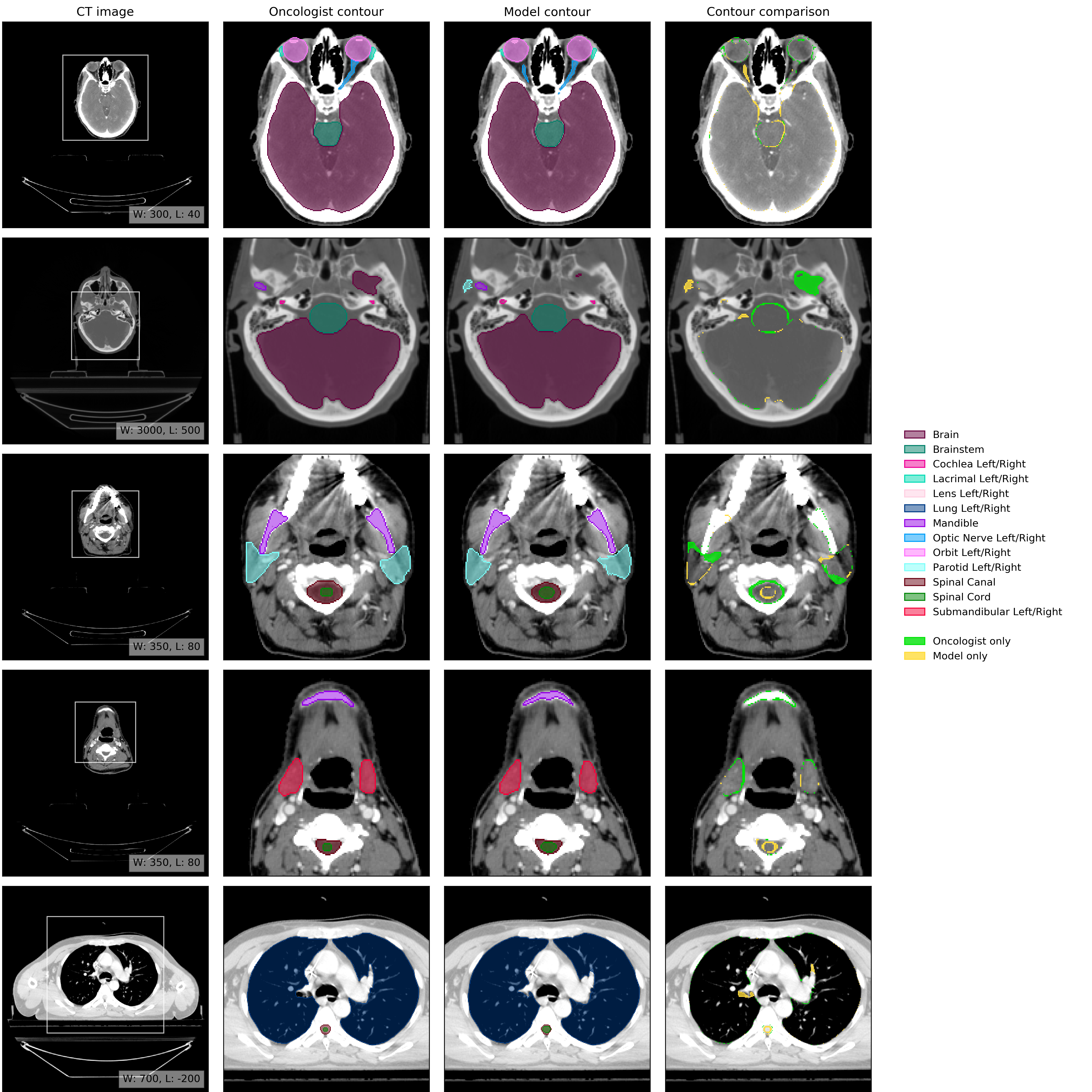}
    \caption{\textbf{Example results}. 
 (\textbf{CT image}) Axial slices at five representative levels from the raw CT scan of a 55-59 year old male patient was selected from the UCLH dataset (patient UCLH-20) were selected to best demonstrate the OARs included in the work. The levels shown as 2D slices have been selected to demonstrate all 21 OARs included in this study. The window levelling has been adjusted for each to best display the anatomy present.
(\textbf{Oncologist contour}) The ground truth segmentation, as defined by experienced radiographers and arbitrated by a head and neck specialist oncologist.
(\textbf{Model contour}) Segmentations produced by our model.
(\textbf{Contour comparison}) Contoured by Oncologist only (green region) or Model only (yellow region).
Two further randomly selected UCLH set scans are shown in \autoref{fig:examples_uclh_2} and \autoref{fig:examples_uclh_3}.
Best viewed on a display.}
    \label{fig:examples_uclh}
\end{figure}

An example of model performance is shown in \autoref{fig:examples_uclh}. We compare our performance (model vs oncologist) to radiographer performance (radiographer vs oncologist). For more information on dataset selection, inclusion and exclusion criteria for patients and OARs please refer to the Methods section. 

\subsection{A New Metric for Assessing Clinical Performance}
In routine clinical care, algorithm-derived segmentation would be reviewed and potentially corrected by a human expert, just as those created by radiographers currently are. Segmentation performance is thus best assessed by determining the fraction of the surface that needs to be redrawn.
The standard volumetric Dice similarity coefficient (volumetric DSC; \cite{Dice1945-mq}) is not well suited to this because it weighs all regions of misplaced delineation equally and independently of their distance from the surface.
For example, two inaccurate segmentations could have a similar volumetric DSC score if one were to deviate from the correct surface boundary by a small amount in many places while another had a large deviation at a single point.
Correcting the former would likely take a considerable amount of time as it would require redrawing almost all of the boundary while the latter could be corrected much faster, potentially with a single edit action.

\begin{figure}[htbp]
    \centering
    \includegraphics[width=0.8\textwidth]{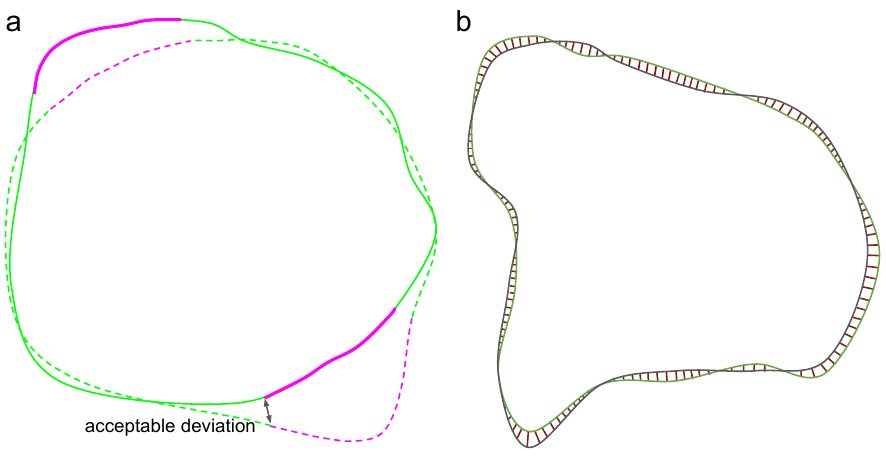}
    \caption{
    \textbf{Surface DSC performance metric}. 
    (\textbf{a}) Illustration of the computation of the surface DSC.
    Continuous line: predicted surface.
    Dashed line: ground truth surface.
    Black arrow: the maximum margin of deviation which may be tolerated without penalty, hereafter referred to by $\tau$.
    Note that in our use case each OAR has an independently calculated value for $\tau$.
    Green: acceptable surface parts (distance between surfaces $\leq\tau$).
    Pink: unacceptable regions of the surfaces (distance between surfaces $>\tau$).
    The proposed surface DSC metric reports the good surface parts compared to the total surface (sum of predicted surface area and ground truth surface area).
    (\textbf{b}) Illustration of the determination of the organ-specific tolerance. 
    Green: segmentation of an organ by oncologist A.
    Black: segmentation by oncologist B.
    Red: distances between the surfaces.
    We defined the organ-specific tolerance as the 95th percentile of the distances collected across multiple segmentations from a subset of seven TCIA scans, where each segmentation was performed a radiographer and then arbitrated by an oncologist, neither of whom had seen the scan previously.}
    \label{fig:surface_dice}
\end{figure}

For quantitative analysis we therefore introduce a new segmentation performance metric, "surface Dice similarity coefficient" (surface DSC) (\autoref{fig:surface_dice}), which assesses the overlap of two surfaces (at a specified tolerance)
instead of the overlap of two volumes. This provides a measure of agreement between the surfaces of two structures, which is where most of the human effort in correcting is usually expended.
In doing so, we also address the volumetric DSC's bias towards large OARs, where the large (and mostly trivial) internal volume accounts for a much larger proportion of the score.

When evaluating the surface DSC we must define a threshold within which variation is clinically acceptable. To do this we first defined organ-specific tolerances (in mm) as a parameter of the proposed metric, $\tau$. We computed these acceptable tolerances for each organ by measuring the inter-observer variation in segmentations between three different consultant oncologists (each with over 10 years experience in OAR delineation) on the validation subset of TCIA images.

To penalise both false negative and false positive parts of the predicted surface, our proposed metrics measures both of the non-symmetric distances between the surfaces and then normalises by the combined surface area. Like the volumetric DSC, the surface DSC ranges from 0 (no overlap) to 1 (perfect overlap).

This means that approximately 95\% of the surface was properly outlined (i.e. within $\tau$ mm of the correct boundary) while 5\% needs to be corrected. There is no consensus as to what constitutes non-significant variation in such segmentation. We thus selected a surface DSC of 0.95 - a stringency which likely far exceeds expert oncologist intra-rater concordance \cite{Daisne2013-uk,HONG2004S157}. For a more formal definition and implementation, please refer to the Methods section. 

\subsection{Model Performance}

Model performance was evaluated alongside that of therapeutic radiographers (each with at least 4 years of experience) segmenting the test set of UCLH images independently of the oncologist-reviewed scans (which we used as our ground truth).

\begin{figure}[htbp]
    \centering
    \includegraphics[width=0.9\textwidth]{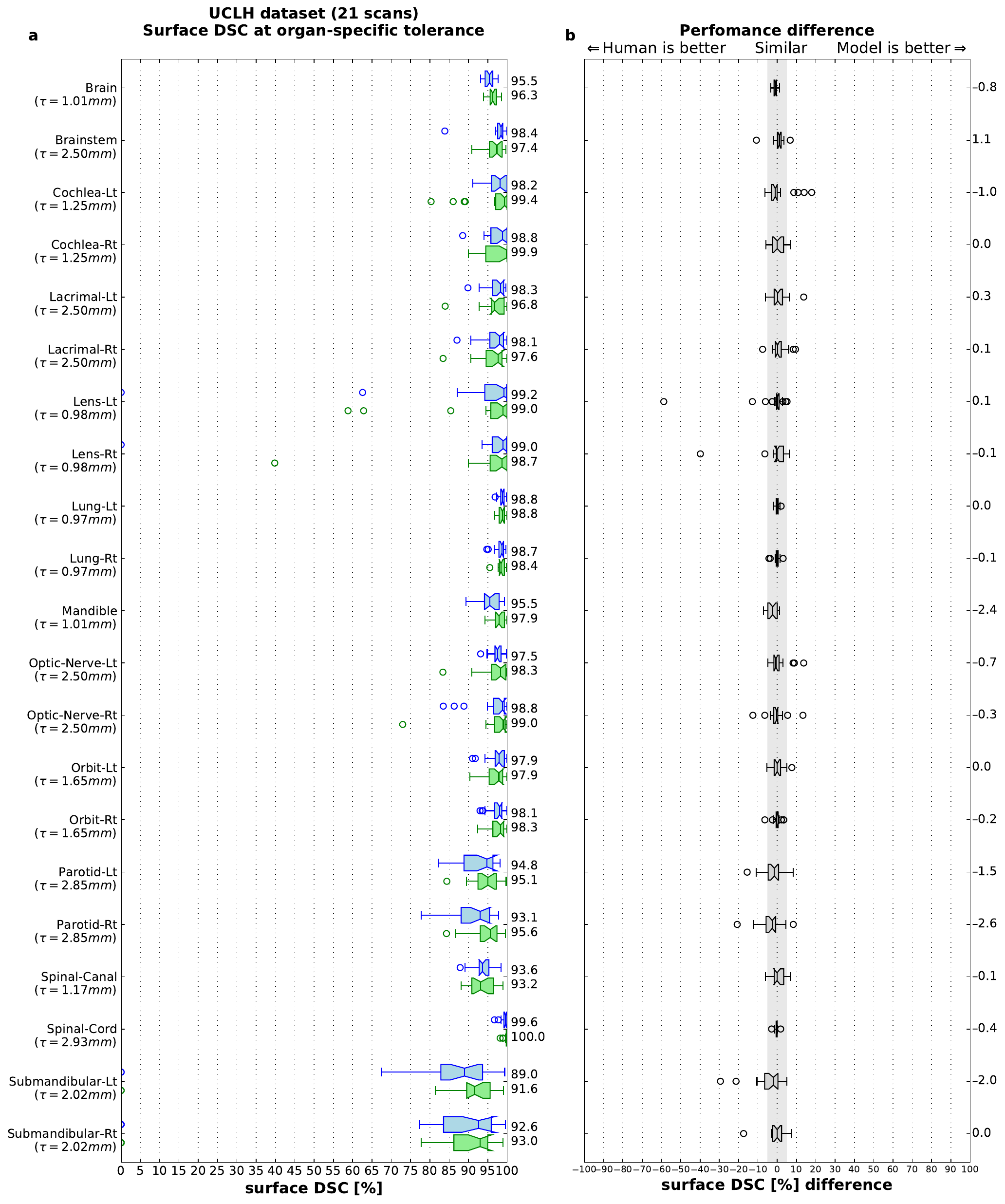}
    \caption{\textbf{UCLH test set: Quantitative performance of the model in comparison to radiographers.} (\textbf{a}) The model achieves a surface DSC similar to humans in all 21 organs at risk (on the UCLH held out test set) when compared to the gold standard for each organ at an organ-specific tolerance $\tau$. Blue: our model, green: radiographers. (\textbf{b}) Performance difference between the model and the radiographers. Each blue dot represents a model-radiographer pair. The grey area highlights non-substantial differences (-5\% to +5\%).\\ 
    The box extends from the lower to upper quartile values of the data, with a line at the median. The whiskers indicate most extreme, non-outlier data points. Where data lies outside 1.5 $\times$ interquartile range it is represented as a circular flier. The notches represent the 95\% confidence interval (CI) around the median.}
    \label{fig:barplot_UCLH}
\end{figure}

The model performed similarly to humans: on all OARs studied there was no clinically meaningful difference between the deep learning model's segmentations and those of the radiographers (\autoref{fig:barplot_UCLH} and \autoref{tab:surface_dice_uclh}). 

\begin{figure}[htbp]
    \centering
    \includegraphics[width=0.9\textwidth]{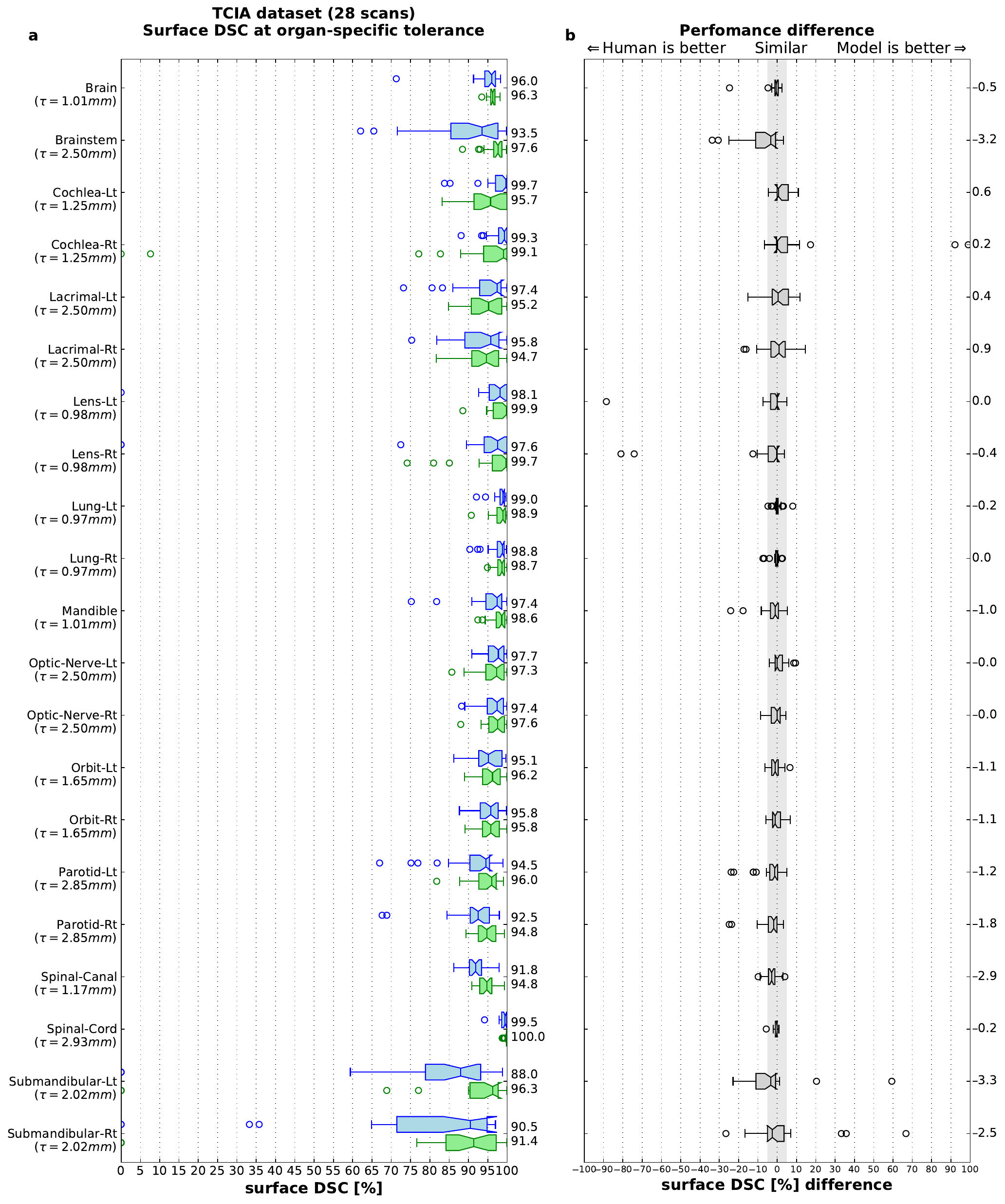}
    \caption{\textbf{Model generalisability to an independent test set from TCIA.} Quantitative performance of the model on the TCIA test set in comparison to radiographers. (\textbf{a}) Surface DSC (on the TCIA open source test set) for the segmentations compared to the gold standard for each organ at an organ-specific tolerance $\tau$. Blue: our model, green: radiographers. (\textbf{b}) Performance difference between the model and the radiographers. Each blue dot represents a model-radiographer pair. Red lines show the mean difference. The grey area highlights non-substantial differences (-5\% to +5\%).\\ 
    The box extends from the lower to upper quartile values of the data, with a line at the median. The whiskers extend from the box to show the range of the data, except where data lies outside 1.5 $\times$ interquartile range, which is represented as a circular flier. The notches represent the 95\% confidence interval (CI) around the median.}
    \label{fig:barplot_TCIA}
\end{figure}

To investigate the generalisability of our model, we additionally evaluate performance on open source scans ('TCIA test set'). These were collected from sites in the USA, where the patient demographics, the clinical pathways for radiotherapy and the scanner type and parameters differed from our UK training set in meaningful ways. Nevertheless, model performance was preserved and, in 19 of 21 OARs, the model performed within the threshold defined for human variability \autoref{fig:barplot_TCIA}. The fact that performance in 2 OARs (brainstem and right lens) was less than that in UK data may relate to issues of image quality in several TCIA test set scans. 

For more detailed results demonstrating surface DSC and volumetric DSC for each individual patient from the TCIA test set please refer to \autoref{tab:surface_dice_tcia} and \autoref{tab:volumetric_dice_tcia} respectively in the appendix. 

\subsection{Comparison to previous work}
An accurate quantitative comparison to previously published literature is difficult due to inherent differences in definitions of ground truth segmentations and varied processes of arbitration and consensus building. Given that the use of surface DSC is novel to this study, we also report the standard volumetric DSC scores achieved by our algorithm (despite the shortcomings of this method) so that direct comparison of our results can be made with those in the existing literature. An overview of past papers which have reported mean volumetric DSC for unedited automatic delineation of head and neck CT OARs can be found in \autoref{tab:literature} and \autoref{tab:literature_full}. Each used different datasets, scanning parameters and labelling protocols, meaning that resulting volumetric DSC results varied significantly. No study other than ours segmented lacrimal glands. We compared these results to those obtained when we applied our model to three different datasets: the TCIA open source test set, an additional test set from the original UCLH dataset (“UCLH test set”) and the dataset released by the Public Domain Database for Computational Anatomy (PDDCA) as part of the 2015 MICCAI head and neck radiotherapy OAR segmentation challenge (“PDDCA test set”; \cite{Raudaschl2017-mn}). To contextualise the performance of our model, radiographer performance is shown on the TCIA test set, and oncologist inter-observer variation is shown on the UCLH test set.

While not the primary test set, we nevertheless present per-patient surface DSC and volumetric DSC for the PDDCA test set in \autoref{tab:surface_dice_pddca} and \autoref{tab:volumetric_dice_pddca} in the appendix.

\begin{table}[htbp]
\label{tab:literature}

\captionabove{\textbf{Volumetric DSC performance of our model and previously published deep-learning models.} An overview of previously published deep-learning based automatic segmentation works that reported volumetric DSC for the OARs included in this study on planning CT scans. Due to the large volume of publications, this overview includes only results of deep learning works. For a full literature overview see \autoref{tab:literature_full}. The datasets and ground truths used varied between studies making comparison difficult. Despite this, we show results alongside our evaluation of our model, radiographers and oncologists against our ground truth across multiple datasets. The latter assesses inter-observer variation between oncologists.}

{
\tiny
\sffamily

\input{table_literature.tex}

\vspace{6pt}
Values for volumetric DCS are mean ($\pm$ standard deviation) unless otherwise stated.
 ``CNN'': convolutional neural network. ``FCN'': fully convolutional network. ``GAN'': generative adversarial network.\\
$^1$ Values estimated from figures; actual values not reported.\\
$^2$ Number of scans per organ varies, see \autoref{tab:num_scans_UCLH}.\\
$^3$ Volumetric DSC estimated from sparse labels. 

}

\end{table}

\section{Discussion}

We demonstrate an automated deep learning-based segmentation algorithm that can perform as well as experienced radiographers for head and neck radiotherapy planning.
Our model was developed using CT scans derived from routine clinical practice, and therefore should be applicable in a hospital setting for segmentation of OARs, routine Radiation Therapy Quality Assurance (RTQA) peer review and reducing the associated variability between different specialists and radiotherapy centres \cite{Wuthrick2015}.

Clinical applicability must be supported not only by a high model performance but also by evidence of model generalisability to new external datasets. To achieve this, we present these results on three separate test sets, one of which (the PDDCA test set) uses a different segmentation protocol.
Here, performance in the majority of OARs was maintained when tested on scans taken from a range of previously unseen international sites.
Although these scans varied in patient demographics, scanning protocol, device manufacturer and image quality, the model still achieved human performance on 19 of the 21 OARs studied; only the right lens and brainstem were below radiographer performance.
For these OARs, the performance of the model might have been lower than expert performance owing to lower image quality.
This is particularly evident for the right lens, where the anatomical borders were quite indistinct in some TCIA test set cases, thus preventing full segmentation by the model (\autoref{fig:hard_lenses}). Moreover, a precise CT definition of the brainstem’s proximal and distal boundaries is lacking, a factor which might have contributed to labelling variability and thus to decreased model performance.
Finally, demographic bias may have resulted from the TCIA data set selecting for cases of more advanced head and neck cancer ~\cite{Bosch2015-tw}, or from variability in the training data~\cite{Nelms2012-dv}.

One major contribution of this article is the presentation of a performance measure that represents the clinical task of OAR correction.  In the first pre-print of this work we introduced surface DSC \cite{nikolov2018deep}, a metric conceived to be sensitive to clinically significant errors in OAR delineation. Surface DSC has recently been shown to be more strongly correlated with the amount of time required to correct a segmentation for clinical use than traditional metrics including volumetric DSC \cite{Vaassen2020evaluation,Kiser2020}. Small deviations in OAR border placement can have a potentially serious impact, increasing the risk of debilitating side effects for the patient. Misplacement by only a small offset may thus require the whole region to be redrawn and in such cases an automated segmentation algorithm may offer no time-savings at all. Volumetric DSC is relatively insensitive to such small changes for large organs as the absolute overlap is also large. Difficulties identifying the exact borders of smaller organs can result in large differences in volumetric DSC even if these differences are not clinically relevant in terms of their effect on radiotherapy treatment. By strongly penalising border placement outside a tolerance determined by consultant oncologists, the surface DSC metric resolves these issues.

While volumetric DSC is therefore not representative of clinical consequences, it remains the most popular metric for evaluating segmentation models and therefore the only metric that allows comparison to previously published works. In recent years, fully convolutional networks became the most popular and successful methodology for OAR segmentation in head and neck CT for de-novo radiotherapy planning \cite{wang2019organ, men2019, tappeiner2019multi, rhee2019automatic, tang2019clinically, van2019deep, jiang2019local, lei2019deepigeos, gao2019focusnet, chan2019convolutional, xue2019shape, wong2020comparing, qiu2020recurrent, liang2020multi, sun2020attentionanatomy}. Although not directly comparable due to different datasets and labelling protocols, our volumetric DSC results compare favourably against the existing published literature for many of the OARs (see \autoref{tab:literature} and \autoref{tab:literature_full} for more details on this and other prior publications). In OARs with inferior volumetric DSC score compared to the published literature, both our model and the human radiographers achieved similar scores. This suggests that current and previously published results are difficult to compare, either due to the inclusion of more difficult cases than previous studies, or due to different segmentation and scanning protocols. To allow more objective comparisons of different segmentation methods, we make our labelled TCIA datasets freely available to the academic community.\footnote{The dataset is available at \url{https://github.com/deepmind/tcia-ct-scan-dataset}.} At least 11 auto-segmentation software solutions are currently available commercially, with varying claims regarding their potential to lower segmentation time during radiotherapy planning \cite{Sharp2014-zq}. The principal factor that determines whether or not automatic segmentation is time-saving during the radiotherapy workflow is the degree to which automated segmentations require correction by oncologists.

The wide variability in state-of-the-art and limited uptake in routine clinical practice motivates the need for clinical studies evaluating model performance in practice. Future work will seek to define the clinical acceptability of the segmented OARs produced by our models, and estimating the time-saving that could be achieved during the radiotherapy planning workflow in a real-world setting.

A number of other study limitations should also be addressed in future work. First, we included only planning CT scans since magnetic resonance imaging (MRI) and Positron Emission Tomography (PET) scans were not routinely performed for all patients in the UCLH dataset. Some OAR classes, such as optic chiasm, require co-registration with MR images for optimal delineation and access to additional imaging has been shown to improve the delineation of optic nerves \cite{Mocnik2018-fv}.
As a result, certain OAR classes were deliberately excluded from this CT-based project and will be addressed in future work which will incorporate MRI scans. A second limitation  regards the classes of OARs in this study. While we present one of the largest sets of reported OARs in the literature
\cite{kosmin2019rapid, tang2019clinically, guo2020organ}, some omissions occurred (e.g., oral cavity) due to an insufficient number of examples in the training data that conformed to a standard international protocol. The number of oncologists used in the creation of our ground truth may not have fully captured the variability in OAR segmentation, or may have been biased towards a particular interpretation of the Brouwer Atlas used as our segmentation protocol. Even in an organ as simple as the spinal cord that is traditionally reliably outlined by auto-segmentation algorithms, there is ambiguity between the inclusion of, for example, the nerve roots. Such variation may widen the thresholds of acceptable deviation in favour of the model despite a consistent protocol. Future work will address these deficits, alongside time-consuming lymph node segmentation.

Finally, neither of the test sets used in this paper include the patients' protected-characteristic status. This is a significant limitation as it  prevents the study of intersectional fairness.

\subsection{Conclusion}

In conclusion, we demonstrate that deep learning can achieve human expert level performance in the segmentation of head and neck OARs in radiotherapy planning CT scans, using a clinically applicable performance metric designed for this clinical scenario. We provide evidence of the generalisability of this model by testing it on patients from different geographies, demographics and scanning protocols. This segmentation algorithm performed with similar accuracy compared to experts and has the potential to improve the speed, efficiency, and consistency of radiotherapy workflows, with an expected positive influence on patient outcomes. Future work will investigate the impact of our segmentation algorithm in clinical practice. 

\section{Methods}
\subsection{Datasets}

University College London Hospitals NHS Foundation Trust (UCLH) serves an urban, mixed socioeconomic and ethnicity population in central London, U.K. and houses a specialist centre for cancer treatment. Data were selected from a retrospective cohort of all adult (>18 years of age) UCLH patients who had computed tomography (CT) scans to plan radical radiotherapy treatment for head and neck cancer between 01/01/2008 and 20/03/2016. Both initial CT images and re-scans were included in the training dataset. Patients with all tumour types, stages and histological grades were considered for inclusion, so long as their CT scans were available in digital form and were of sufficient diagnostic quality. The standard CT pixel spacing was 0.976mm by 0.976mm by 2.5mm, and scans with non-standard spacing (with the exception of 1.25mm spacing scans which were subsampled) were excluded to ensure consistent performance metrics during training. Note that for the TCIA test set, below, the in-plane pixel spacing was not used as an exclusion criteria, i ranged from 0.94mm to 1.27mm. For the PDDCA test set we included all scans, and the voxels varied between 2mm - 3mm in height and 0.98mm - 1.27mm in the axial dimension. The wishes of patients who had requested that their data should not be shared for research were respected. 

Of the 513 patients who underwent radiotherapy at UCLH within the given study dates, a total of 486 patients (838 scans), mean age 57, male 337, female 146, gender unknown 3, met the inclusion criteria. Of note, no scans were excluded on the basis of poor diagnostic quality. Scans from UCLH were split into a training set (389 patients, 663 scans), validation set (51 patients, 100 scans) and test set (46 patients, 75 scans). From the selected test set 19 patients (21 scans) underwent adjudicated Contouring described below. No patient was included in multiple datasets: in cases where multiple scans were present for a single patient, all were included in the same subset. Where multiple scans were present for a single patient, this reflects CT scans taken for the purpose of re-planning radiotherapy due to anatomical changes during a course of treatment. It is important for models to perform well in both scenarios as treatment naive and post-radiotherapy OAR anatomy can differ. However, to avoid potential correlation between the same organs segmented twice in the same dataset, care was taken to avoid this in the TCIA test set (see below).

Twenty-one organs at risk were selected throughout the head and neck area to represent a wide range of anatomical regions.
We used a combination of segmentations sourced from those used clinically at UCLH and additional segmentations performed in-house by trained radiographers.

We divided our UCLH dataset into the following categories:
(1) \textbf{Training set}: Used to train the model, a combination of UCLH clinical segmentations and in-house segmentations, some of which were only 2D slices
\footnote{Due to the time required to segment larger organs manually, we initially relied heavily on sparse segmentations to make efficient use of the radiographers' time.}.
(2) \textbf{UCLH Validation set}: Used to evaluate model performance and steer additional dataset priorities, this used in-house segmentations only, as we didn't want to overfit to any clinical bias.
(3) \textbf{UCLH test set}: Our primary result set, each scan has every OAR labelled and was independently segmented from scratch by two radiographers before one of the pair of scans (chosen arbitrarily) was reviewed and corrected by an experienced radiation oncologist.

As these scans were taken from UCLH patients not present elsewhere, and to consider generalisability, we curated additional open source CT scans available from The Cancer Imaging Archive (TCGA-HNSC and Head-Neck Cetuximab) \cite{Bosch2015-tw, Clark2013-zx, Zuley2016-nb}.
The open source (4) \textbf{TCIA validation set} and (5) \textbf{TCIA test set} were both labelled in the same way as our UCLH test set.

\begin{figure}
    \centering
    \includegraphics[width=\textwidth]{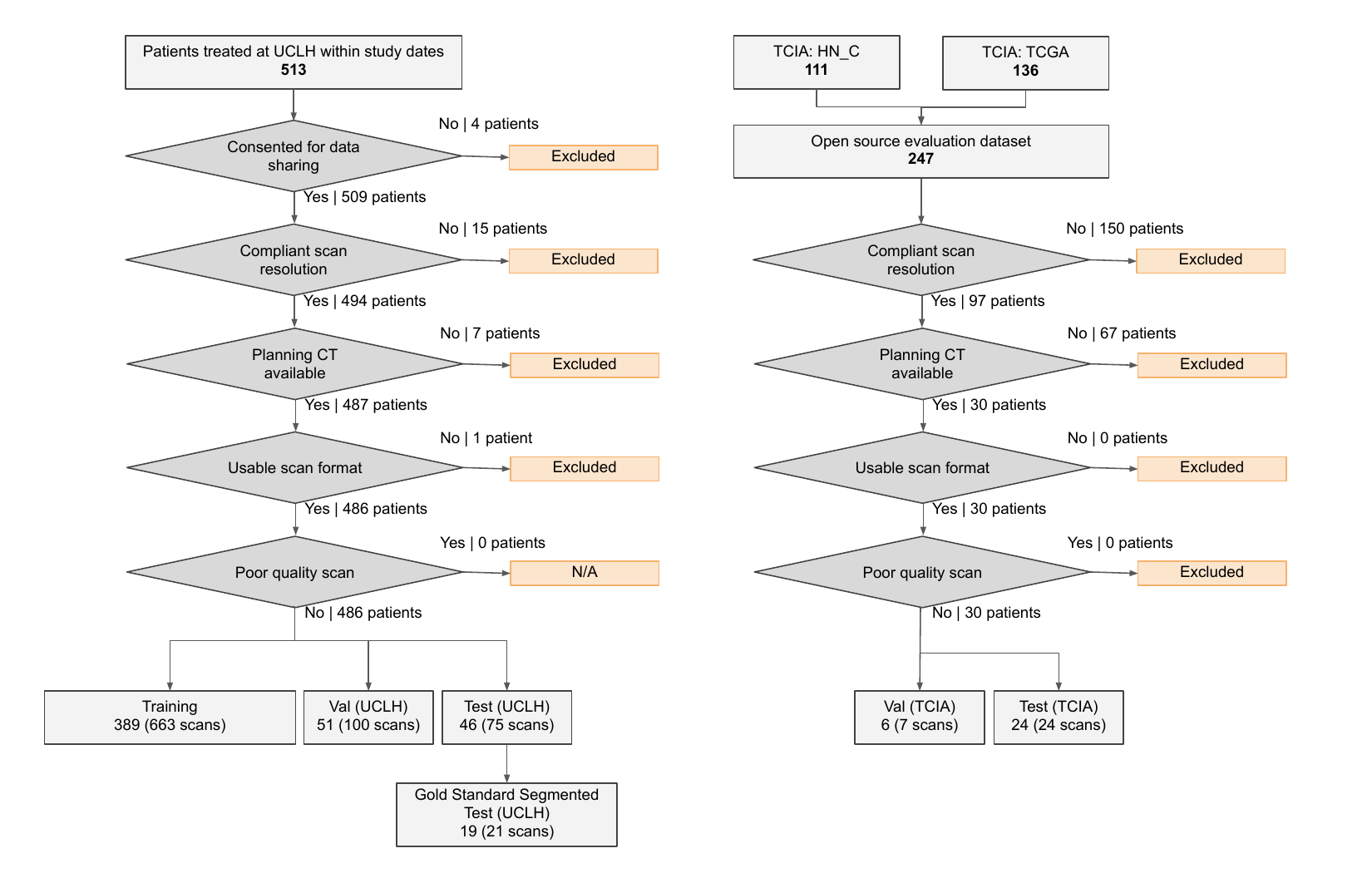}
    \caption{\textbf{Case selection from UCLH and TCIA CT datasets.} A consort-style diagram demonstrating the application of inclusion and exclusion criteria to select the training, validation (val) and test sets used in this work.}
    \label{fig:consort_diagram}
\end{figure}

Non-CT planning scans and those that did not meet the same slice thickness as the UCLH scans (2.5mm) were excluded. These were then manually segmented in-house according to the Brouwer Atlas (\cite{Brouwer2015-qr}; the segmentation procedure is described in further detail below). We included 31 scans (22 Head-Neck Cetuximab, 9 TCGA-HNSC) which met these criteria, which we further split into validation (6 patients, 7 scans) and test (24 patients, 24 scans) sets (\autoref{fig:consort_diagram}). The original segmentations from the Head-Neck Cetuximab dataset were not included; a consensus assessment by experienced radiographers and oncologists found the segmentations either non-conformant to the selected segmentation protocol or below the quality that which would be acceptable for clinical care. The original inclusion criteria for Head-Neck Cetuximab were patients with stage III-IV carcinoma of the oropharynx, larynx, and hypopharynx, having Zubrod performance of 0-1, and meeting predefined blood chemistry criteria between 11/2005 to 03/2009. The TCGA-HNSC dataset included patients treated for Head-Neck Squamous Cell Carcinoma, with no further restrictions being apparent. For more information please refer to the specific citations \cite{Zuley2016-nb,Bosch2015-tw}.

All test sets were kept separate during model training and validation. \autoref{tab:dataset_characteristics} describes in further detail the demographics and characteristics within the datasets; to obtain a balanced demographic in each of the test, validation and training datasets we sampled randomly stratified splits and selected one that minimised the differences between the key demographics in each dataset. 

In addition, the (6) \textbf{PDDCA open source dataset} consisted of 15 patients selected from the Head-Neck Cetuximab open source dataset \cite{Bosch2015-tw}; due to differences in selection criteria and test/validation/training set allocation there were five scans present in both the TCIA and PDDCA test sets. This dataset was used without further post-processing and only accessed once for assessing volumetric DSC performance. The PDDCA test set differ from the TCIA test set in both segmentation protocol and axial slice thickness. For more details on the dataset characteristics and preprocessing please refer to the work of Raudaschl and colleagues \cite{Raudaschl2017-mn}.

\autoref{tab:dataset_characteristics} details the characteristics of these datasets and the patient demographics.

\newcommand{\grayzero}{0}
\begin{table}[htbp]
\captionabove{Dataset Characteristics}
\label{tab:dataset_characteristics}
\centering
\sffamily
\scriptsize
\begin{tabular}{llcccccc}\toprule
                      &                              & \multicolumn{3}{c}{UCLH}         & \multicolumn{2}{c}{TCIA} & \multicolumn{1}{c}{PDDCA}   \\
\cmidrule(lr){3-5}
\cmidrule(lr){6-7}
\cmidrule(lr){8-8}
                       &                              & Train     & Validation & Test    & Validation & Test    & Test    \\
\midrule
\multicolumn{2}{l}{Total scans (patients)}		&663 (389)	&100 (51)	&21 (19)	&7 (6)	&24 (24)	&15 (15)	\\
\multicolumn{2}{l}{Average patient age}		&57.1	&57.5	&59.6	&56.5	&59.9	&58.6	\\
								\\
Sex	&Female	&207 (115)	&36 (19)	&7 (6)	&2 (2)	&2 (2)	&2 (2)	\\
	&Male	&450 (271)	&64 (32)	&14 (13)	&5 (4)	&20 (20)	&9 (9)	\\
	&Unknown	&6 (3)	&\grayzero	&\grayzero	&\grayzero	&2 (2)	&4 (4)	\\
								\\
Tumour site	&Oropharynx	&145 (86)	&27 (15)	&7 (6)	&\grayzero	&8 (8)	&2 (2)	\\
	&Lip, oral cavity and pharynx	&80 (52)	&20 (8)	&4 (4)	&1 (1)	&3 (3)	&\grayzero	\\
	&Tongue	&53 (26)	&8 (5)	&1 (1)	&2 (2)	&7 (7)	&\grayzero	\\
	&Larynx	&46 (31)	&8 (3)	&2 (2)	&2 (2)	&4 (4)	&\grayzero	\\
	&Nasopharynx	&48 (24)	&5 (3)	&\grayzero	&\grayzero	&\grayzero	&\grayzero	\\
	&Head, face and neck	&37 (23)	&8 (3)	&1 (1)	&\grayzero	&\grayzero	&\grayzero	\\
	&Nasal Cavity	&32 (19)	&2 (1)	&1 (1)	&\grayzero	&\grayzero	&\grayzero	\\
	&Connective and soft tissue	&37 (18)	&2 (1)	&1 (1)	&\grayzero	&\grayzero	&\grayzero	\\
	&Hypopharynx	&17 (10)	&1 (1)	&\grayzero	&2 (1)	&1 (1)	&\grayzero	\\
	&Accessory sinus	&10 (7)	&2 (1)	&\grayzero	&\grayzero	&\grayzero	&\grayzero	\\
	&Oesophagus	&6 (2)	&1 (1)	&\grayzero	&\grayzero	&\grayzero	&\grayzero	\\
	&Other	&33 (20)	&\grayzero	&\grayzero	&\grayzero	&1 (1)	&\grayzero	\\
	&Unknown	&119 (71)	&16 (9)	&4 (3)	&\grayzero	&\grayzero	&13 (13)	\\
								\\
Source	&TCGA	&-	&-	&-	&2 (2)	&7 (7)	&\grayzero	\\
	&HN\_Cetux	&-	&-	&-	&5 (4)	&17 (17)	&15 (15)	\\
								\\
Site	&UCLH	&663 (389)	&100 (51)	&21 (19)	&\grayzero	&\grayzero	&\grayzero	\\
	&MD Anderson Cancer Clinic	&\grayzero	&\grayzero	&\grayzero	&2 (2)	&7 (7)	&\grayzero	\\
	&Unknown (US)	&\grayzero	&\grayzero	&\grayzero	&5 (4)	&17 (17)	&15 (15)	\\
\bottomrule
\end{tabular}\\[6pt]
Tumour sites are derived from ICD codes.
Numbers show number of scans with the number of unique patients in parenthesis.
"TCGA": The Cancer Genome Atlas Head-Neck Squamous Cell Carcinoma\cite{Zuley2016-nb}, an open source dataset hosted on TCIA.
"HN\_Cetux": Head-Neck Cetuximab, an open source dataset hosted on TCIA\cite{Bosch2015-tw}.
"PDDCA": Public Domain Database for Computational Anatomy dataset released as part of the 2015 challenge in the segmentation of head and neck anatomy at the International Conference On Medical Image Computing \& Computer Assisted Intervention (MICCAI).
\end{table}

\subsection{Clinical taxonomy}
In order to select which OARs to include in the study, we used the Brouwer Atlas (consensus guidelines for delineating OARs for head and neck radiotherapy, defined by an international panel of radiation oncologists; \cite{Brouwer2015-qr}). From this, we excluded those regions which required additional magnetic resonance imaging for segmentation, were not relevant to routine head and neck radiotherapy, or that were not used clinically at UCLH. This resulted in a set of 21 organs at risk; see \autoref{tab:organs}.

\subsection{Clinical labelling \& annotation}
Due to the large variability of segmentation protocols used and annotation quality in the UCLH dataset, all segmentations from all scans selected for inclusion in the training set were manually reviewed by a radiographer with at least 4 years experience in the segmentation of head and neck OARs. Volumes that did not conform to the Brouwer Atlas were excluded from training. In order to increase the number of training examples, additional axial slices were randomly selected for further manual OAR segmentations to be added based on model performance or perceived imbalances in the dataset. These were then produced by a radiographer with at least 4 years experience in head and neck radiotherapy, arbitrated by a second radiographer with the same level of experience. The total number of examples from the original UCLH segmentations and the additional slices added are provided in \autoref{tab:organs}.

\begin{table}[htbp]
\captionabove{Taxonomy of segmentation regions. }
\label{tab:organs}
\centering
\sffamily
\scriptsize

\input{table_organs.tex}

\end{table}

\begin{figure}
    \centering
    \includegraphics[width=\textwidth]{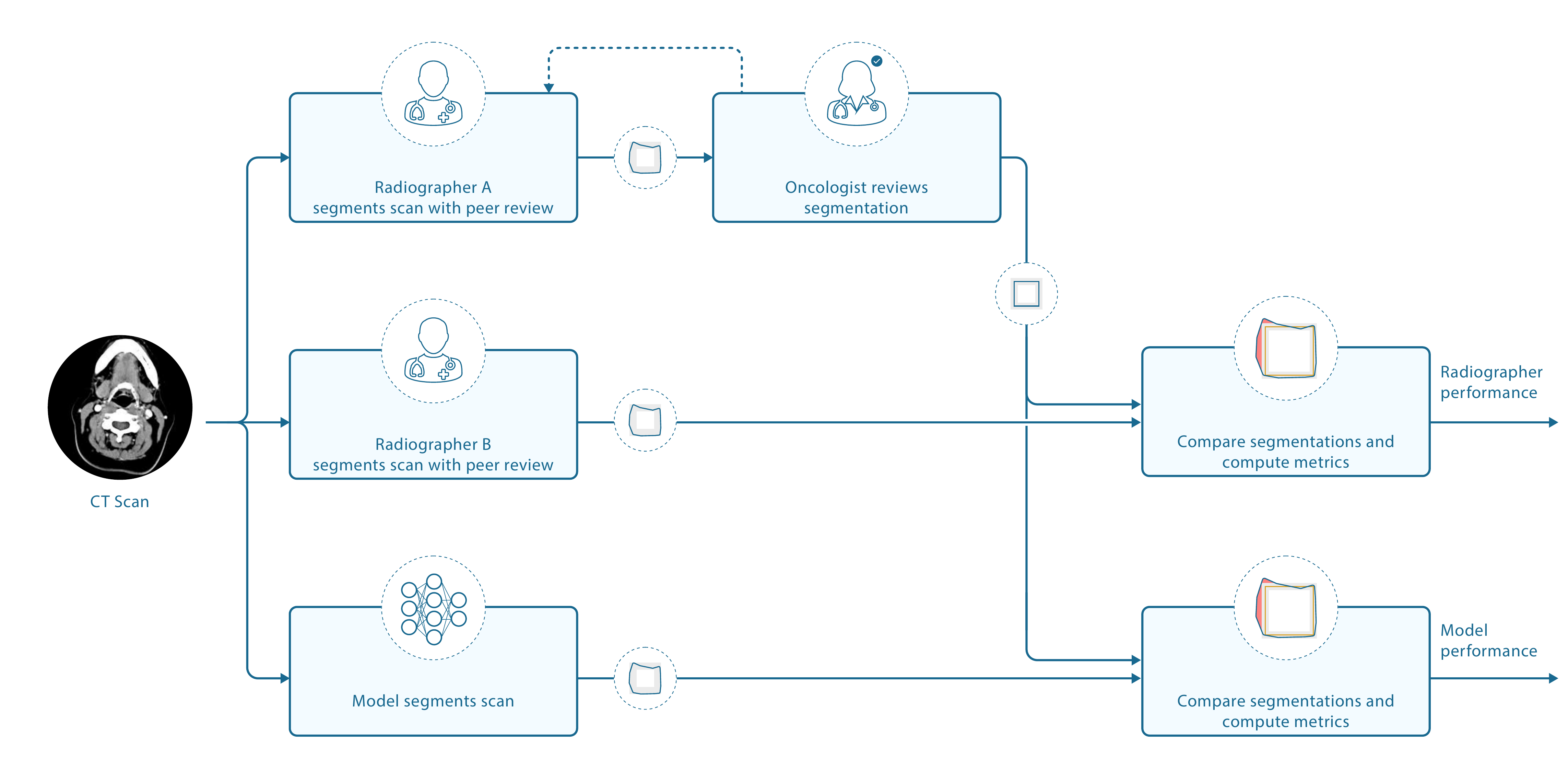}
    \caption{\textbf{Process for segmentation of ground truth and radiographer OAR volumes}. The flowchart illustrates how the ground truth segmentations were created and compared with independent radiographer segmentations and the model. For the ground truth each CT scan in the TCIA test set was segmented first by a radiographer and peer reviewed by a second radiographer. This then went through one or more iterations of review and editing with a specialist oncologist before creating a ground truth used to compare with the segmentations produced by both the model and additional radiographers.}
    \label{fig:segmentation_process}
\end{figure}

For the TCIA test and validation sets, the original dense segmentations were not used due to poor adherence to the chosen study protocol. To produce the ground truth labels, the full volumes of all 21 OARs included in the study were segmented. This was done initially by a radiographer with at least four years experience in the segmentation of head and neck OARs and then arbitrated by a second radiographer with similar experience. Further arbitration was then performed by a radiation oncologist with at least five years post-certification experience in head and neck radiotherapy. The same process was repeated with two additional radiographers working independently but after peer arbitration these segmentations were not reviewed by an oncologist; rather they became the human reference to which the model was compared. This is shown schematically in \autoref{fig:segmentation_process}. Prior to participation all radiographers and oncologists were required to study the Brouwer Atlas for head and neck OAR segmentation \cite{Brouwer2015-qr} and demonstrate competence in adhering to these guidelines.

\subsection{Model architecture}
\begin{figure}
    \centering
    \includegraphics[width=\textwidth]{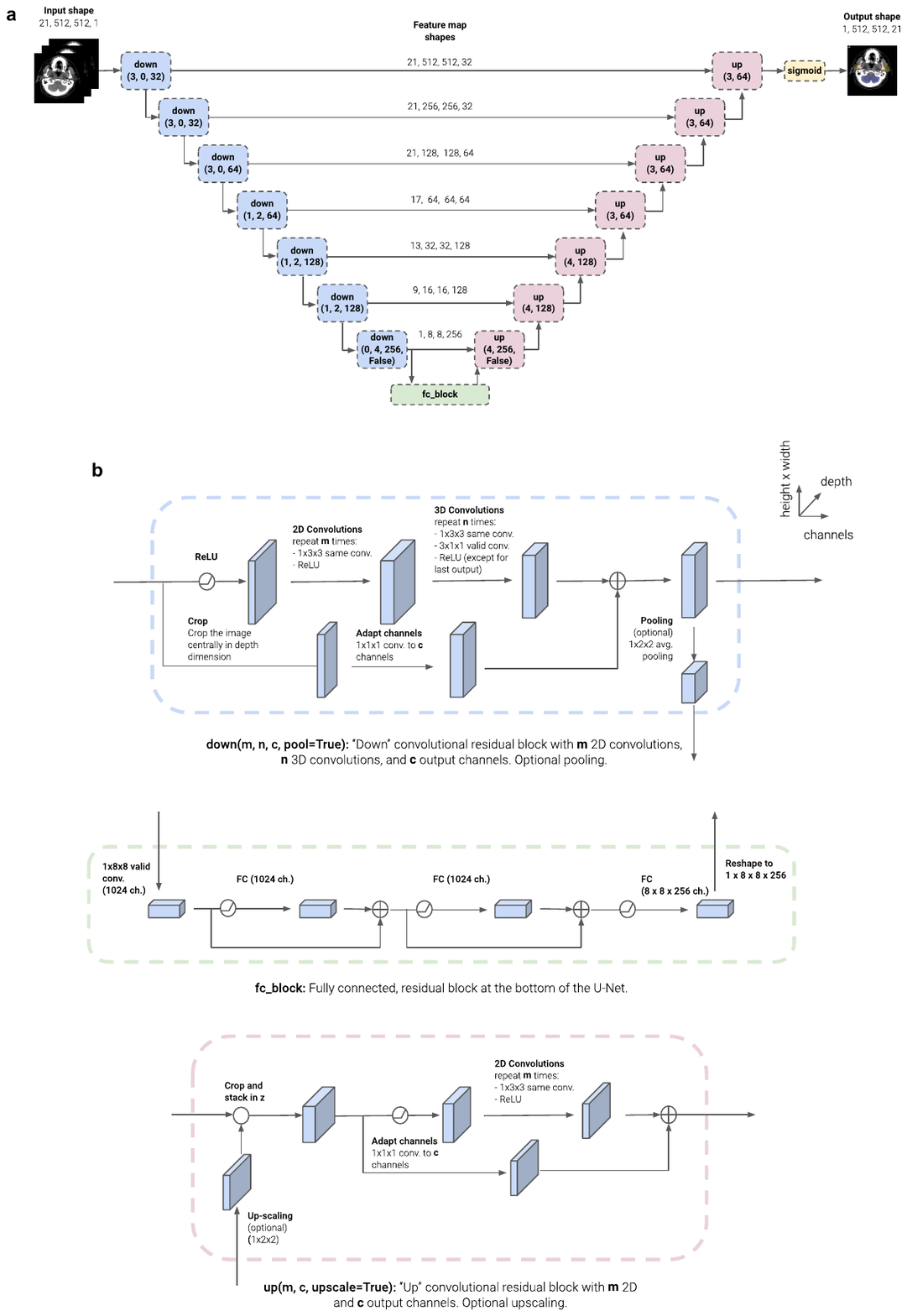}
    \caption{\textbf{3D U-Net model architecture}. (\textbf{a}) At training time, the model receives 21 contiguous CT slices, which are processed through a series of “down” blocks, a fully connected block, and a series of “up” blocks to create a segmentation prediction. (\textbf{b}) A detailed view of the convolutional residual down and up blocks, and the residual fully connected block.}
    \label{fig:model_architecture}
\end{figure}

We used a residual 3D U-Net architecture with 8 levels (see \autoref{fig:model_architecture}). Our network takes in a CT volume (single channel) and outputs a segmentation mask with 21 channels, where each channel contains the binary segmentation mask for a specific OAR. The network consists of 7 residual convolutional blocks in the downward path, a residual fully connected block at the bottom, and 7 residual convolutional blocks in the upward path. A 1x1x1 convolution layer with sigmoidal activation produces the final output in the original resolution of the input image. Each predicted slice has 21 slices of context\footnote{The 21 slices context (i.e. 21$\times$2.5mm = 52.5mm) were found to provide the optimal context. It has nothing to do with the 21 OARs used in this study.}.

We trained our network with a regularised top-k-percent pixel-wise binary cross-entropy loss \cite{Wu2016-yt}: for each output channel, the top-k loss selects only the k\% most difficult pixels (those with the highest binary cross-entropy), and only adds their contribution to the total loss. This speeds up training and helps the network to tackle the large class imbalance and to focus on difficult examples. 

We regularised the model using standard L2 weight regularisation with scale $10^{-6}$ and extensive data augmentation: we used random in-plane (i.e. in x- and y- directions only) translation, rotation, scaling, shearing, mirroring, elastic deformations, and pixel-wise noise. We used uniform translations between -32 and 32 pixels; uniform rotations between -9 and 9 degrees; uniform scaling factors between 0.8 and 1.2; and uniform shear factors between -0.1 and 0.1. We mirrored images (and adjusted corresponding left and right labels) with a probability of 0.5. We performed elastic deformations by placing random displacement vectors (standard deviation: 5mm, in-plane displacements only) on a control point grid with 100mm x 100mm x 100mm spacing and by deriving the dense deformation field using cubic b-spline interpolation. In the implementation all spatial transformations are first combined to a dense deformation field, which is then applied to the image using bilinear interpolation and extrapolation with zero padding.  We added zero mean Gaussian intensity noise independently to each pixel with a standard deviation of 20 Hounsfield Units. 

We trained the model with the Adam optimiser \cite{Kingma2014-rm} for 120,000 steps and a batch size of 32 (32 GPUs) using synchronous SGD. We used an initial learning rate of $10^{-4}$ and scaled the learning rate by 1/2, 1/8, 1/64, and 1/256 at timesteps 24,000, 60,000, 108,000, and 114,000, respectively.

We used the validation set to select the model which performed at over 95\% for the most OARs according to our chosen surface DSC performance metric, breaking ties by preferring better performance on more clinically impactful OARs and the absolute performance obtained.

\subsection{Performance metrics}
\begin{figure}[htbp]
    \centering
    \includegraphics{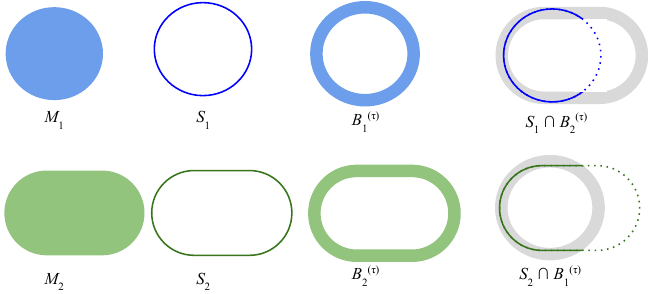}
    \caption{Illustrations of masks, surfaces, border regions, and the ``overlapping'' surface at tolerance $\tau$}
    \label{fig:examples}
\end{figure}

All performance metrics are reported for each organ independently (e.g. separately for just the left parotid), so we only need to deal with binary masks (e.g. a left parotid voxel and a non left-parotid voxel). 
Masks are defined as a subset of $\R^3$, i.e. $\mathcal{M} \subset \R^3$ (see \autoref{fig:examples}).

The volume of a mask is denoted as $\abs{\cdot}$, with
\begin{align*}
    \abs{\mathcal{M}} &= 
    \int\limits_{\mathcal{M}} d\vec{x} \,.
\end{align*}
With this notation the standard (volumetric) DSC for two given masks $\mathcal{M}_1$ and $\mathcal{M}_2$ can be written as
\begin{align*}
    C_{\text{DSC}} = \frac{
    2\abs{\mathcal{M}_1 \cap \mathcal{M}_2}}
    {\abs{\mathcal{M}_1} + \abs{\mathcal{M}_2}} \,.
\end{align*}
In the case of sparse ground truth segmentations (i.e. only a few slices of the CT scan are labelled), we estimate the volumetric DSC by aggregating data from labelled voxels across multiple scans and patients as
\begin{align*}
    C_{\text{DSC, est}} = \frac{
    2\sum_p \abs{
    \mathcal{M}_{1,p} 
    \cap \mathcal{M}_{2,p}
    \cap \mathcal{L}_p}
    }{
    \sum_p \abs{\mathcal{M}_{1,p} \cap \mathcal{L}_p} 
    +
    \abs{\mathcal{M}_{2,p} \cap \mathcal{L}_p}
    } \,,
\end{align*}
where the mask $\mathcal{M}_{1,p}$ and the labelled region $\mathcal{L}_p$ represent the sparse ground truth segmentation for a patient $p$ and the mask $\mathcal{M}_{2,p}$ is the full volume predicted segmentation for patient $p$.

Due to the shortcomings of the volumetric DSC metric for the presented radiotherapy use case, we introduce the ``surface DSC'' metric, which assesses the overlap of two surfaces (at a specified tolerance) instead of the overlap of two volumes (see Results section). A surface is the border of a mask, $\mathcal{S} = \partial \mathcal{M}$, the area of a surface is denoted as
\begin{align*}
    \abs{\mathcal{S}} &= 
    \int\limits_{\mathcal{S}} d\vec{\sigma} 
\end{align*}
where $\vec{\sigma} \in \mathcal{S}$ is a point on the surface, using an arbitrary parameterisation. The mapping from this parameterisation to a point in $\R^3$ is denoted as $\vec{\xi} : \mathcal{S} \rightarrow \R^3$, i.e. $\vec{x} = \vec{\xi}(\vec{\sigma})$. With this we can define the border region $\mathcal{B}_i^{(\tau)} \subset \R^3$, for the surface $\mathcal{S}_i$, at a given tolerance $\tau$ as
\begin{align*}
    \mathcal{B}_i^{(\tau)} &= 
    \left\lbrace
    \vec{x} \in \R^3 
    \mid
    \exists \vec{\sigma} \in \mathcal{S}_i, 
    \norm{\vec{x} - \vec{\xi}(\vec{\sigma})} \leq \tau
    \right\rbrace\,,
\end{align*}
(see \autoref{fig:examples} for an example). Using these definitions we can write the ``surface DSC at tolerance $\tau$'' as
\begin{align*}
    R_{i,j}^{(\tau)} &= \frac{
    \abs{\mathcal{S}_i \cap \mathcal{B}_j^{(\tau)}}
    +
    \abs{\mathcal{S}_j \cap \mathcal{B}_i^{(\tau)}}
    }{
    \abs{\mathcal{S}_i}
    +
    \abs{\mathcal{S}_j}
    }\,,
\end{align*}
using an informal notation for the intersection of the surface with the boundary, i.e.:
\begin{align*}
    \abs{\mathcal{S}_i \cap \mathcal{B}_j^{(\tau)}}
    :=
    \int\limits_{\mathcal{S}_i} 
    \mathds{1}_{\mathcal{B}_j^{(\tau)}}
    \left(
    \vec{\xi}(\vec{\sigma})
    \right)
    d\vec{\sigma}
\end{align*}


\subsection{Implementation of surface DSC}
The computation of surface integrals on sampled images is not straightforward, especially for medical images, where the voxel spacing is usually not equal in all three dimensions. The common approximation of the integral by counting surface voxels can lead to substantial systematic errors. 

Another common challenge is the representation of the surface with voxels. As the surface of a binary mask is located between voxels, a definition of ``surface voxels'' in the raster-space of the image introduces a bias: using foreground voxels to represent the surface leads to an underestimation of the surface, while the use of background voxels leads to an overestimation. 

\begin{figure}[htbp]
    \centering
    \includegraphics{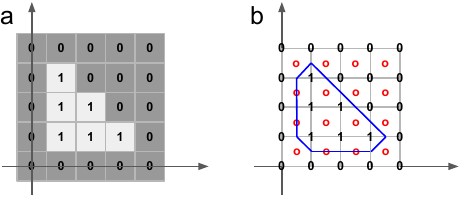}
    \caption{\textbf{2D illustration of the implementation of the surface DSC}. (\textbf{a}) A binary mask displayed as image. The origin of the image raster is (0,0). (\textbf{b}) The surface points (red circles) are located in a raster that is shifted half of the raster spacing on each axis. Each surface point has 4 neighbours in 2D (8 neighbours in 3D). The local contour (blue line) assigned to each surface point (red circle) depends on the neighbour constellation.}
    \label{fig:surface_dice_implementation}
\end{figure}

Our proposed implementation uses a surface representation that provides less biased estimates but still allows us to compute the performance metrics with linear complexity ($\mathcal{O}(N)$, with $N:$ number of voxels). We place the surface points between the voxels on a raster that is shifted by half of the raster spacing on each axis (see \autoref{fig:surface_dice_implementation} for a 2D illustration).
For 3D images, each point in this raster has 8 neighbouring voxels. As we analyse binary masks, there are only $2^8 = 256$ possible neighbour constellations. For each of these constellations we compute the resulting triangles using the marching cube triangulation \cite{Lorensen1987-qz} and store the surface area of the triangles (in mm$^2$) in a look-up table. With this look-up table we then create a surface image (on the above mentioned raster) that contains zeros at positions that have 8 identical neighbours or the local surface area at all positions that have both foreground and background neighbours.
These surface images are created for the masks $\mathcal{M}_1$ and $\mathcal{M}_2$. 
Additionally we create a distance map from each of these surface images using the distance transform algorithm \cite{Felzenszwalb2012-uy}. Iterating over the non-zero elements in the first surface image and looking up the distance from the other surface in the corresponding distance map allows to create a list of tuples (surface element area, distance from other surface).
From this list we can easily compute the surface area by summing up the area of the surface elements that are within the tolerance. 
To account for the quantised distances -- there is only a discrete set $\mathcal{D} = \left\{ \sqrt{(n_1 d_1)^2 + (n_2 d_2)^2 + (n_3 d_3)^2} \mid n_1, n_2, n_3 \in \mathds{N} \right\}$ of distances between voxels in a 3D raster with spacing  $(d_1,d_2,d_3)$  -- we also round the tolerance to the nearest neighbour in the set $\mathcal{D}$ for each image before computing the surface DSC.
For more details, please have a look at our open source implementation of the surface DSC, available from \url{https://github.com/deepmind/surface-distance}.

\section{Code availability}
The codebase for the deep learning framework makes use of proprietary components and we are unable to publicly release this code. However, all experiments and implementation details are described in sufficient detail in the methods section to allow independent replication with non-proprietary libraries. The surface DSC performance metric code is available at \url{https://github.com/deepmind/surface-distance}. 

\section{Data availability}
The clinical data used for training and validation sets were collected and de-identified at University College London Hospitals NHS Foundation Trust. Data were used with both local and national permissions. They are not publicly available and restrictions apply to their use. The data, or a subset, may be available from UCLH NHS Foundation Trust subject to local and national ethical approvals. The released test/validation set data was collected from two datasets hosted on The Cancer Imaging Archive (TCIA). The subset used, along with the ground truth segmentations added is available at \url{https://github.com/deepmind/tcia-ct-scan-dataset}.

\section{Acknowledgement}
We thank the patients treated at UCLH whose scans were used in the work, A.~Zisserman, D.~King, D.~Barrett, V.~Cornelius, C.~Beltran, J.~Cornebise, R.~Sharma, J.~Ashburner, J.~Good and N.~Haji for discussions, M.~Kosmin for his review of the published literature, J.~Adler for discussion and review of the manuscript, A.~Warry, U.~Johnson, V.~Rompokos and the rest of the UCLH Radiotherapy Physics team for work on the data collection, R.~West for work on the visuals, C.~Game, D.~Mitchell and M.~Johnson for infrastructure and systems administration, A.~Paine at Softwire for engineering support at UCLH, A.~Kitchener and the UCLH Information Governance team for support, J.~Besley and M.~Bawn for legal assistance, K.~Ayoub, K.~Sullivan and R.~Ahmed for initiating and supporting the collaboration, the DeepMind Radiographer Consortium made up of B.~Garie, Y.~McQuinlan, K.~Hampton, S.~Ireland, K.~Fuller, H.~Frank, C.~Tully, A.~Jones and L.~Turner, and the rest of the DeepMind team for their support, ideas and encouragement.

G.R. and H.M. were supported by University College London and the National Institute for Health Research (NIHR) University College London Hospitals Biomedical Research Centre. The views expressed are those of the author(s) and not necessarily those of the NHS, the NIHR or the Department of Health. 

\section{Author contributions}
M.S., T.B., O.R., J.L., R.M., H.M., S.A.M., D.D'S., C.C., \& K.S. initiated the project \newline
S.B., R.M., D.C., C.B., D.D'S., C.C. \& J.L., created the dataset \newline
S.B, S.N., J.D.F., A.Z., Y.P., C.H., H.A. \& O.R. contributed to software engineering \newline
S.N., J.D.F., B.R.P. \& O.R. designed the model architectures \newline
D.R.C. manually segmented the images \newline
R.M., D.C., C.B., D.D'S., S.A.M., H.M., G.R., C.H., A.K. \& J.L. contributed clinical expertise \newline
C.M., J.L., T.B., S.A.M., K.S. \& O.R. managed the project \newline
C.H., C.K., M.L., J.L., S.N., S.B., J.D.F., H.M., G.R. \& O.R. wrote the paper \newline

\section{Competing financial interests}
G.R., H.M. and the D.R.C. were paid contractors of DeepMind and/or Google Health. The authors have no other competing interests to disclose.

\addcontentsline{toc}{section}{References}

\bibliographystyle{ieeetran}
\bibliography{radical}

\pagebreak
\clearpage
\pagenumbering{arabic}
\renewcommand*{\thepage}{A\arabic{page}}
\appendix

\section{Appendix}

\begin{table}[htbp]
\captionabove{Surface DSC on TCIA data set}
\label{tab:surface_dice_tcia}

{
\tiny
\sffamily

\hspace{-20mm}
\input{table_surface_dice_TCIA.tex}

\vspace{6pt}
Numbers below the organ name show the average surface area of this organ in the test set.\\
M: our model performance\\
H: human (radiographer) performance\\
numbers in brackets indicate that this organ for this patient would not be segmented in current clinical practise\\
$^1$: aggregated only over organs that would be segmented for this patient in current clinical practise. I.e. numbers in brackets were excluded. \\
Colours indicate the performance difference:\\

\begin{tabular}{ll}
\cellcolor{magenta!50} < -10\% &(model is worse) \\
\cellcolor{magenta!25} -10\% to -5\%& (model is slightly worse) \\
-5\% -- +5\% &(model and human are on par) \\
\cellcolor{green!25} +5\% to +10\% &(model is slightly better)\\
\cellcolor{green!50} > +10\% &(model is better)
\end{tabular}
}
\end{table}

\begin{table}[htbp]
\captionabove{Volumetric DSC on TCIA data set}
\label{tab:volumetric_dice_tcia}
{
\tiny
\sffamily
\hspace{-20mm}
\input{table_volumetric_dice_TCIA.tex}
\\[6pt]
Numbers below the organ name show the average volume of this organ in the test set.\\
M: our model performance\\
H: human (radiographer) performance\\
Colors indicate performance differences: green: model is better, red: model is worse

}
\end{table}

\begin{table}[htbp]
\captionabove{Surface DSC on PDDCA data set}
\label{tab:surface_dice_pddca}
\centering
{
\tiny
\sffamily
\input{table_surface_dice_pddca.tex}
\\[6pt]
Numbers below the organ name show the average surface area of this organ in the PDDCA test set.\\

Colours indicate the performance difference:\\

\begin{tabular}{ll}
\cellcolor{magenta!50} < -10\% &(model is worse) \\
\cellcolor{magenta!25} -10\% to -5\%& (model is slightly worse) \\
-5\% -- +5\% &(model and human are on par) \\
\cellcolor{green!25} +5\% to +10\% &(model is slightly better)\\
\cellcolor{green!50} > +10\% &(model is better)
\end{tabular}
}
\end{table}

\begin{table}[htbp]
\captionabove{Volumetric DSC on PDDCA data set}
\label{tab:volumetric_dice_pddca}
\centering
{
\tiny
\sffamily
\input{table_volumetric_dice_pddca.tex}
\\[6pt]
Numbers below the organ name show the average volume of this organ in the PDDCA test set.\\

Colours indicate the performance difference:\\

\begin{tabular}{ll}
\cellcolor{magenta!50} < -10\% &(model is worse) \\
\cellcolor{magenta!25} -10\% to -5\%& (model is slightly worse) \\
-5\% -- +5\% &(model and human are on par) \\
\cellcolor{green!25} +5\% to +10\% &(model is slightly better)\\
\cellcolor{green!50} > +10\% &(model is better)
\end{tabular}
}
\end{table}

\begin{table}[htbp]
\captionabove{Surface DSC on UCLH data set}
\label{tab:surface_dice_uclh}
\centering
{
\tiny
\sffamily
\input{table_surface_dice_UCLH.tex}
\\[6pt]
Numbers below the organ name show the average surface area of this organ in the UCLH test set.\\

M: our model performance\\
H: human (radiographer) performance\\

Colours indicate the performance difference:\\

\begin{tabular}{ll}
\cellcolor{magenta!50} < -10\% &(model is worse) \\
\cellcolor{magenta!25} -10\% to -5\%& (model is slightly worse) \\
-5\% -- +5\% &(model and human are on par) \\
\cellcolor{green!25} +5\% to +10\% &(model is slightly better)\\
\cellcolor{green!50} > +10\% &(model is better)
\end{tabular}
}
\end{table}

\begin{table}[htbp]
\captionabove{Volumetric DSC on UCLH data set}
\label{tab:volumetric_dice_uclh}
\centering
{
\tiny
\sffamily
\input{table_volumetric_dice_UCLH.tex}
\\[6pt]
Numbers below the organ name show the average volume of this organ in the UCLH test set.\\

M: our model performance\\
H: human (radiographer) performance\\

Colors indicate performance differences: green: model is better, red: model is worse
}
\end{table}

\begin{table}[htbp]
\captionabove{Number of labelled scans in UCLH test set}
\label{tab:num_scans_UCLH}
{
\tiny
\sffamily
\hspace{-15mm}
\input{table_num_scans_UCLH.tex}
}
\end{table}

\begin{figure}[htbp]
    \centering
    \includegraphics[width=\textwidth]{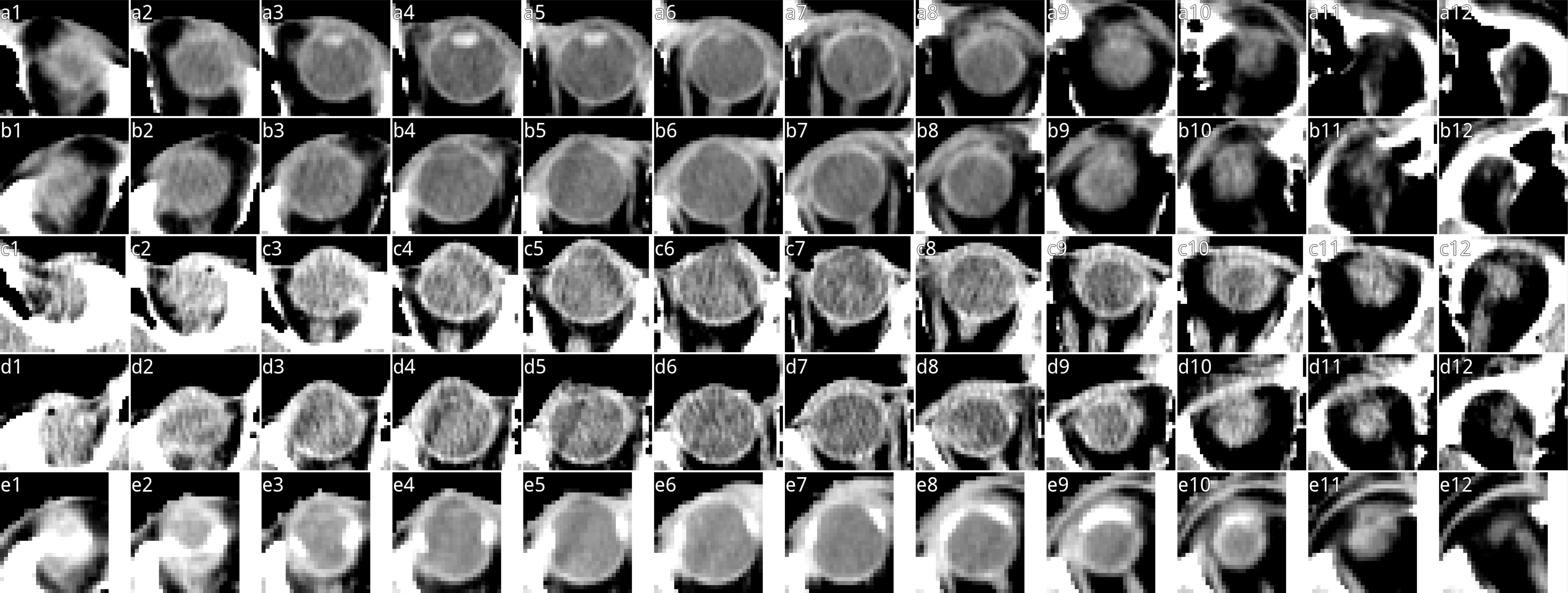}
    \caption{\textbf{Missed lens predictions across the TCIA test set}. Consecutive axial slices of eyes showing both a typical lens and the four cases where the model predictions omitted the lens. The window level is at a constant W:140 L:0. (\textbf{a1-a12}) 12 slices through a single eye in which the model was able to detect the lens, which is clearly visible in (a3-a6). (a1) is the upper most slice, (a12) the lower most. (\textbf{b1-e12})~Similar to the first row, but these four cases are those for which the model was unable to differentiate the lens from the rest of the eye. Note that all four cases are considerably more challenging than for the first row.
    }
    \label{fig:hard_lenses}
\end{figure}

\clearpage

\begin{figure}[htbp]
    \centering
    \includegraphics[width=\textwidth]{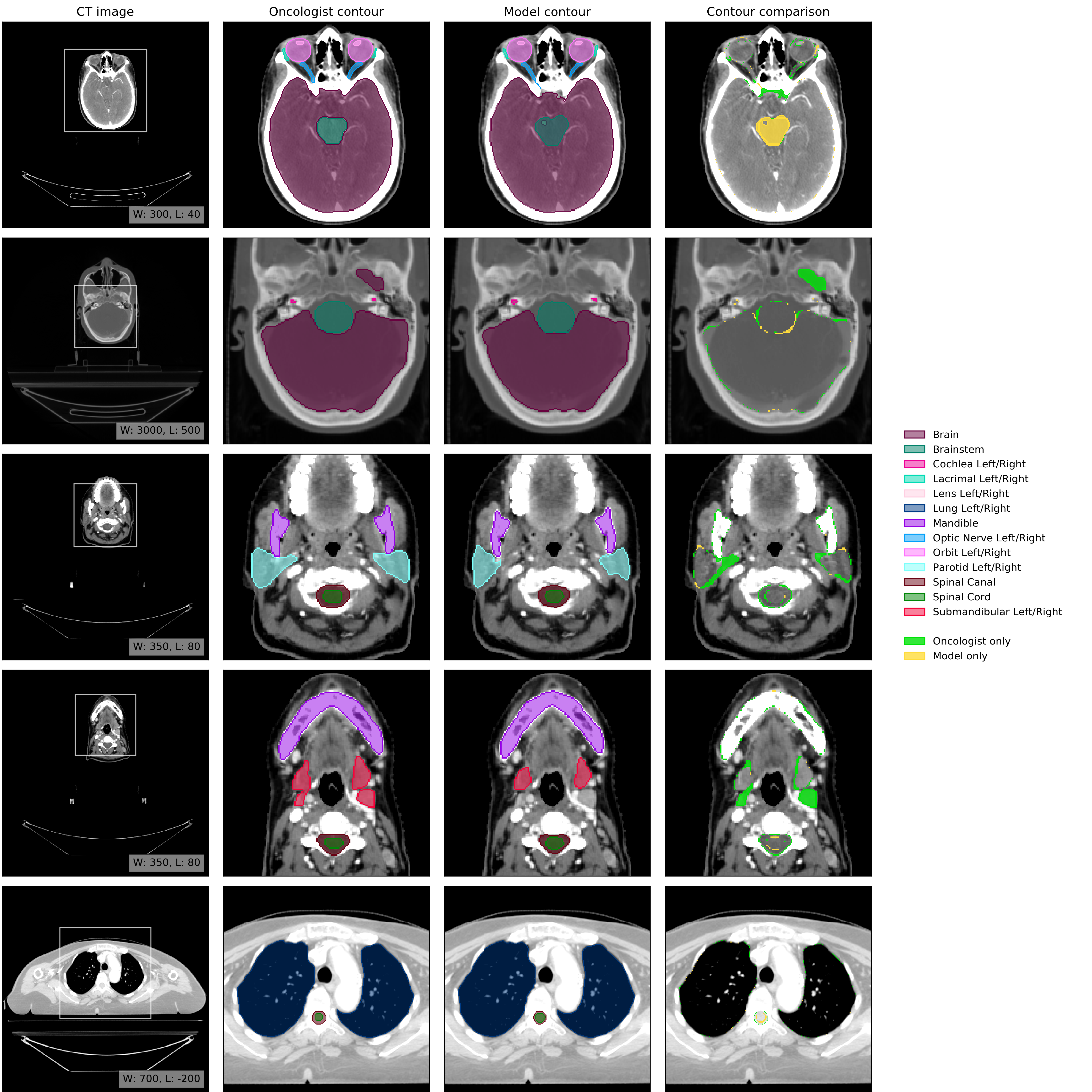}
    \caption{\textbf{Example results.} Axial slices at five representative levels from the raw CT scan of 70-74 year old female patient from the UCLH test set. The levels shown as 2D slices have been selected to demonstrate all 21 OARs included in this study. The window levelling has been adjusted for each to best display the anatomy present.
(\textbf{Oncologist contour}) The ground truth segmentation, as defined by experienced radiographers and arbitrated by a head and neck specialist oncologist.
(\textbf{Model contour}) Segmentations produced by our model.
(\textbf{Contour comparison}) Contoured by Oncologist only (green region) or Model only (yellow region).
Two further randomly selected UCLH set scans are shown in \autoref{fig:examples_uclh_2} and \autoref{fig:examples_uclh_3}.
Best viewed on a display.}
    \label{fig:examples_uclh_2}
\end{figure}

\clearpage
\begin{figure}[htbp]
    \centering
    \includegraphics[width=\textwidth]{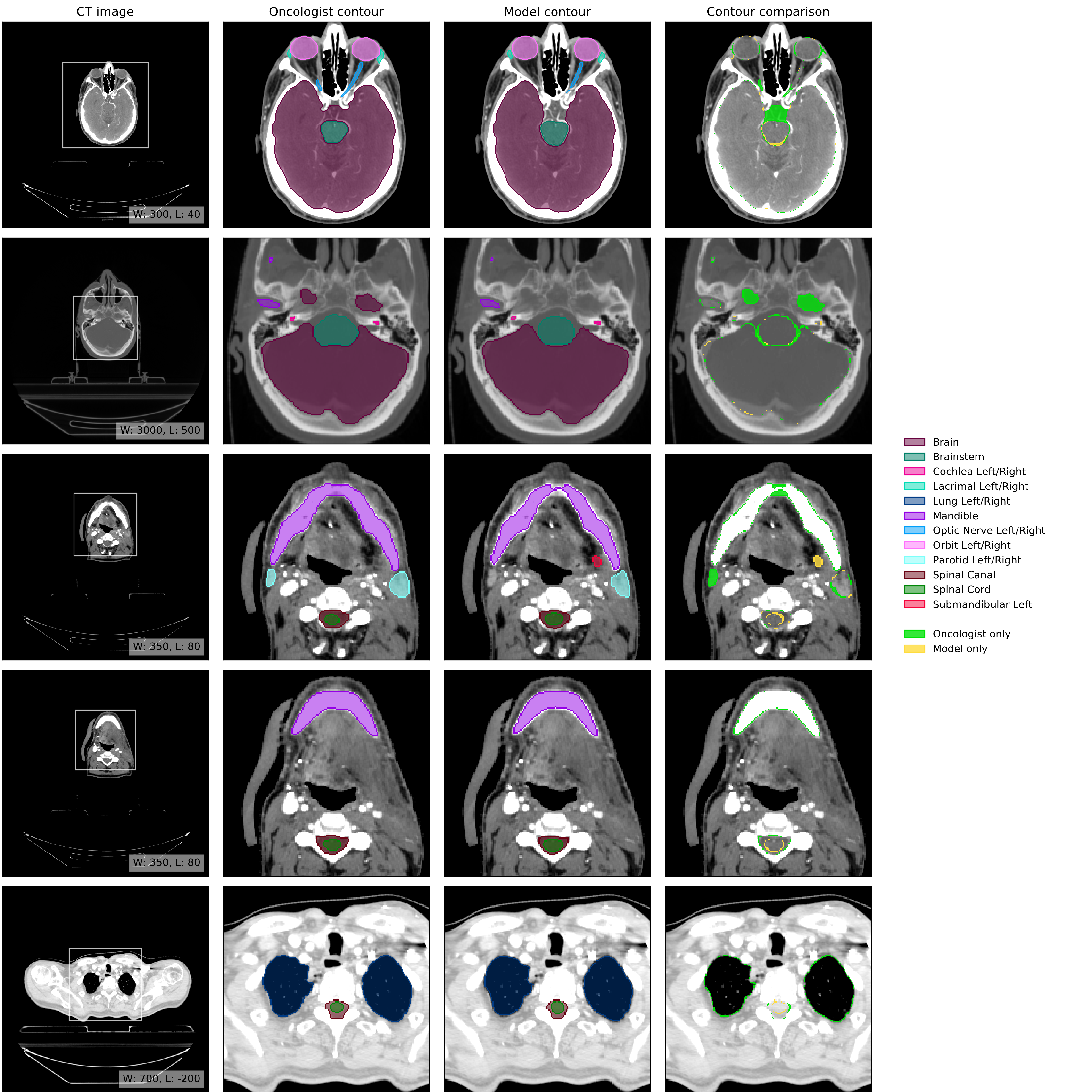}
    \caption{\textbf{Example results.} Axial slices at five representative levels from the raw CT scan of 70-74 year old male patient from the UCLH test set. The levels shown as 2D slices have been selected to demonstrate all 21 OARs included in this study. The window levelling has been adjusted for each to best display the anatomy present.
(\textbf{Oncologist contour}) The ground truth segmentation, as defined by experienced radiographers and arbitrated by a head and neck specialist oncologist.
(\textbf{Model contour}) Segmentations produced by our model.
(\textbf{Contour comparison}) Contoured by Oncologist only (green region) or Model only (yellow region).
Two further randomly selected UCLH set scans are shown in \autoref{fig:examples_uclh_2} and \autoref{fig:examples_uclh_3}.
Best viewed on a display.}
    \label{fig:examples_uclh_3}
\end{figure}

\begin{figure}[htbp]
    \centering
    \includegraphics[width=\textwidth]{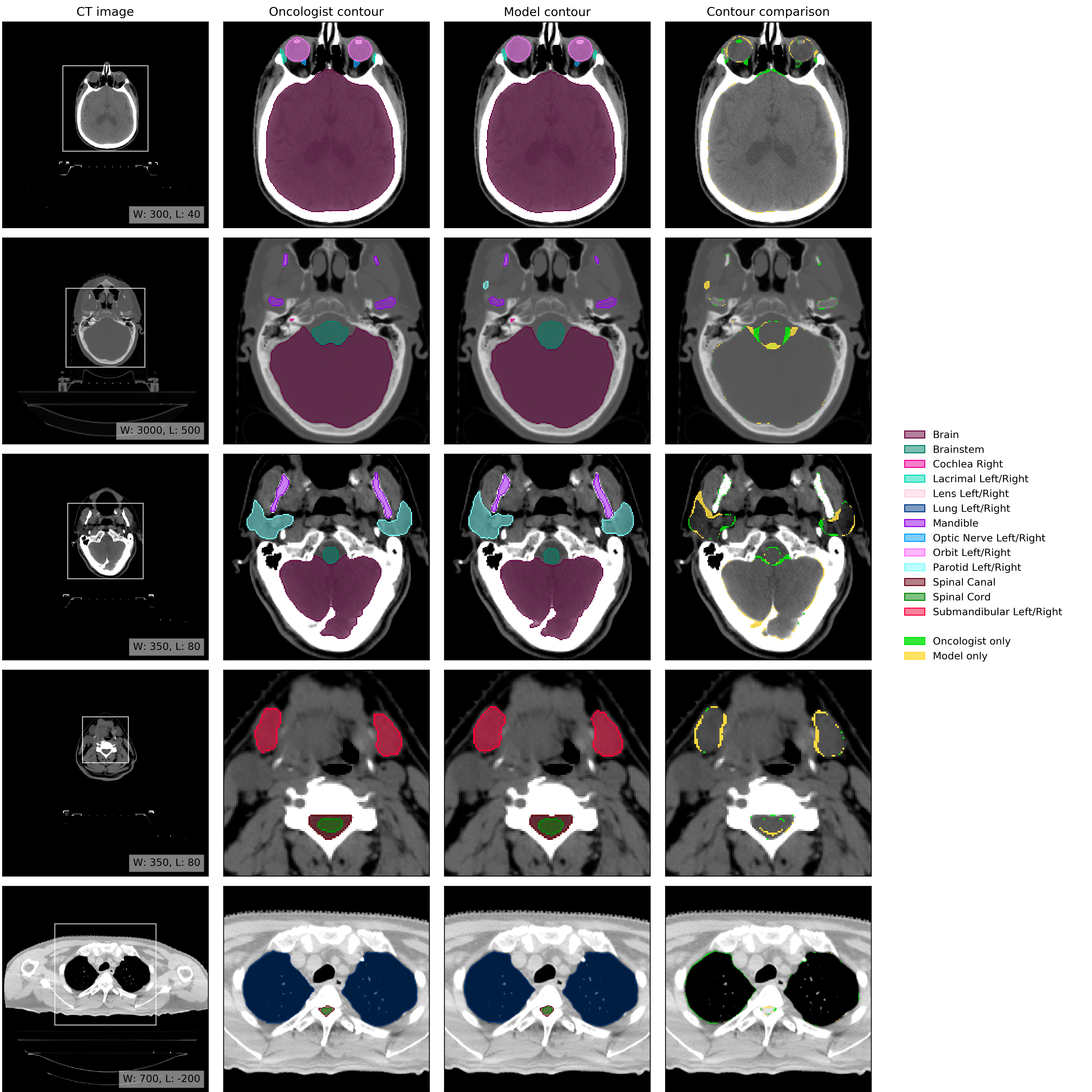}
    \caption{\textbf{Example results}. 
 (\textbf{a1-e1}) Axial slices at five representative levels from the raw CT scan of a 66 year old male patient with a right base of tongue cancer and bilateral lymph node involvement selected from the Head-Neck Cetuximab TCIA dataset (patient 0522c0057; \cite{Bosch2015-tw}) were selected to best demonstrate the OARs included in the work. The levels shown as 2D slices have been selected to demonstrate all 21 OARs included in this study. The window levelling has been adjusted for each to best display the anatomy present.
(\textbf{a2-e2}) The ground truth segmentation, as defined by experienced radiographers and arbitrated by a head and neck specialist oncologist.
(\textbf{a3-e3}) Segmentations produced by our model.
(\textbf{a4-e4}) Overlap between the model (yellow line) and the ground truth (blue line).
Two further randomly selected TCIA set scans are shown in \autoref{fig:examples_tcia_2} and \autoref{fig:examples_tcia_3}.
Best viewed on a display.}
    \label{fig:examples_tcia}
\end{figure}
\begin{figure}[htbp]
    \centering
    \includegraphics[width=\textwidth]{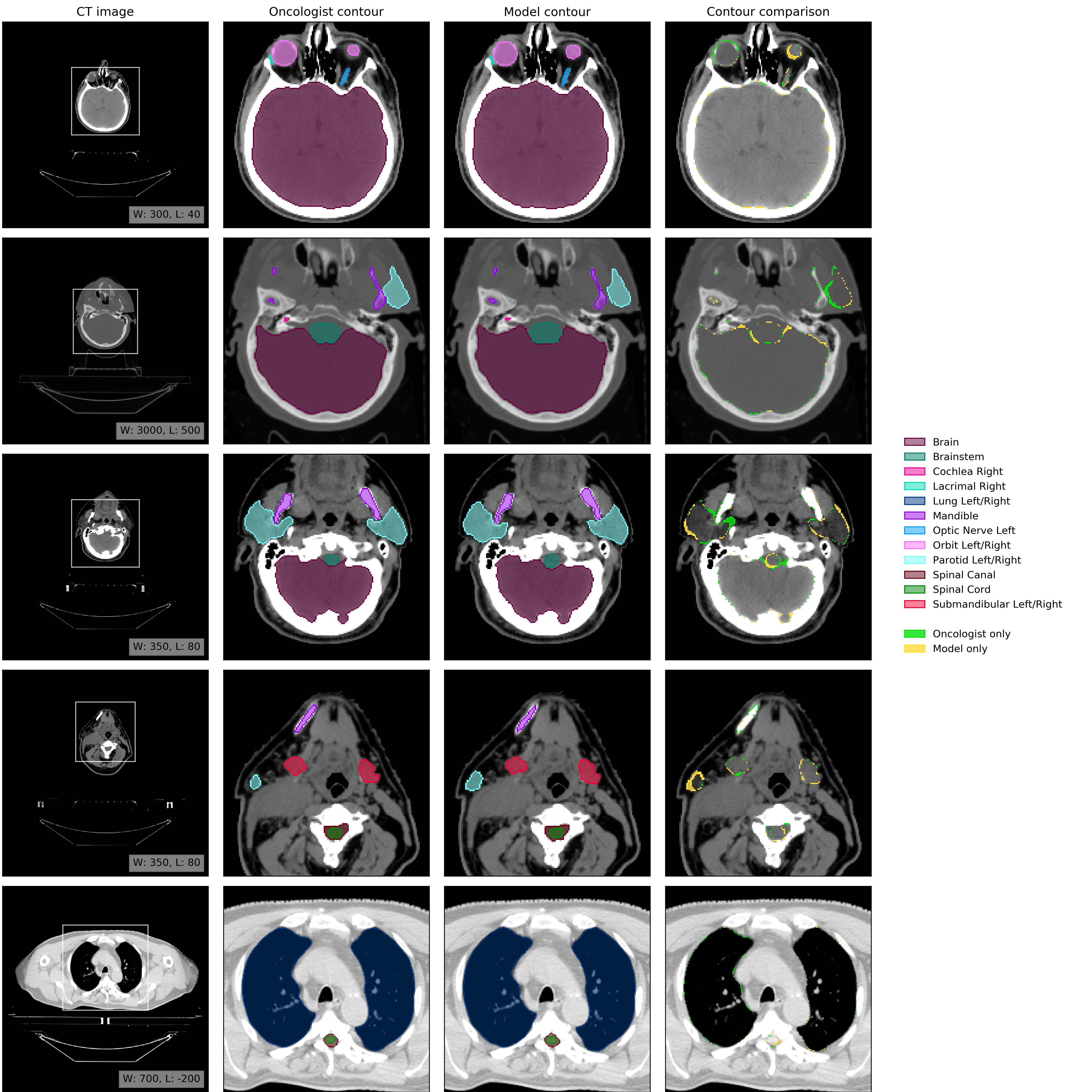}
    \caption{\textbf{Example results from a second randomly selected case from the  TCIA test set}. Five axial slices from the scan of a 58 year old male patient with a cancer of the right tonsil selected from the Head-Neck Cetuximab TCIA dataset (patient 0522c0416; \cite{Bosch2015-tw}). (a1-e1) The raw CT scan slices at five representative levels were selected to best demonstrate the OARs included in the work. The window levelling has been adjusted for each to best display the anatomy present. (a2-e2) The ground truth segmentation was defined by experienced radiographers and arbitrated by a head and neck specialist oncologist. (a3-e3) The model produced segmentations of the same structures. Overlap between the model (yellow line) and the ground truth (blue line) is shown in (a4-e4). Best viewed on a display.}
    \label{fig:examples_tcia_2}
\end{figure}

\clearpage
\begin{figure}[htbp]
    \centering
    \includegraphics[width=\textwidth]{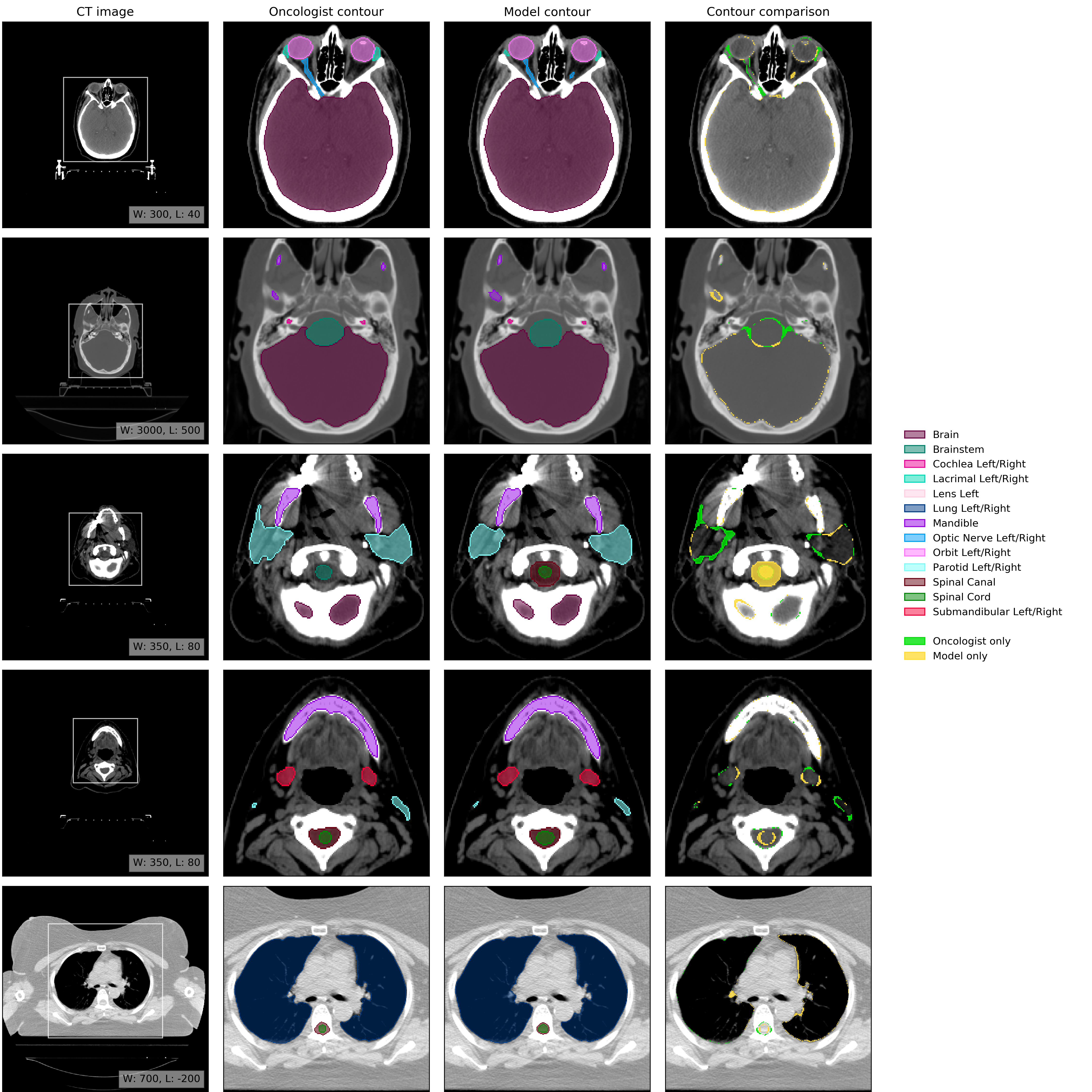}
    \caption{\textbf{Example results from a third randomly selected case from the TCIA test set}. Five axial slices from the scan of a 53 year old female patient with a left oropharyngeal cancer with base of tongue invasion included selected from the Head-Neck Cetuximab TCIA dataset (patient 0522c0251; \cite{Bosch2015-tw}). (a1-e1) The raw CT scan slices at five representative levels were selected to best demonstrate the OARs included in the work. The window levelling has been adjusted for each to best display the anatomy present. (a2-e2) The ground truth segmentation was defined by experienced radiographers and arbitrated by a head and neck specialist oncologist. (a3-e3) The model produced segmentations of the same structures. Overlap between the model (yellow line) and the ground truth (blue line) is shown in (a4-e4). Best viewed on a display.}
    \label{fig:examples_tcia_3}
\end{figure}

\begin{table}[htbp]
\label{tab:literature_full}
\captionabove{\textbf{Volumetric DSC performance of our model and previously published results.} An overview of previously published automatic segmentation works that reported volumetric DSC for the OARs included in this study on planning head and neck CT scans. The datasets and ground truths used varied between studies making comparison difficult. Despite this, we show results alongside our evaluation of our model, radiographers and oncologists against our ground truth across multiple datasets. The latter assesses inter-observer variation between oncologists.}

\tiny
\sffamily
\input{table_literature_full.tex}

\vspace{6pt}
Values for volumetric DCS are mean ($\pm$ standard deviation) unless otherwise stated. ``ABAS'': atlas based auto segmentation. ``CNN'': convolutional neural network. ``FCN'': fully convolutional network. ``GAN'': generative adversarial network. ``HAS'':  hybrid atlas-based segmentation.\\

$^1$ merged brainstem and spinal cord.
$^2$ Values estimated from figures; actual values not reported.
$^3$ Median; mean not reported.
$^4$ Number of scans per organ varies, see \autoref{tab:num_scans_UCLH}.
$^5$ Volumetric DSC estimated from sparse labels.
\end{table}

\begin{figure}[htbp]
    \centering
    \includegraphics[width=\textwidth]{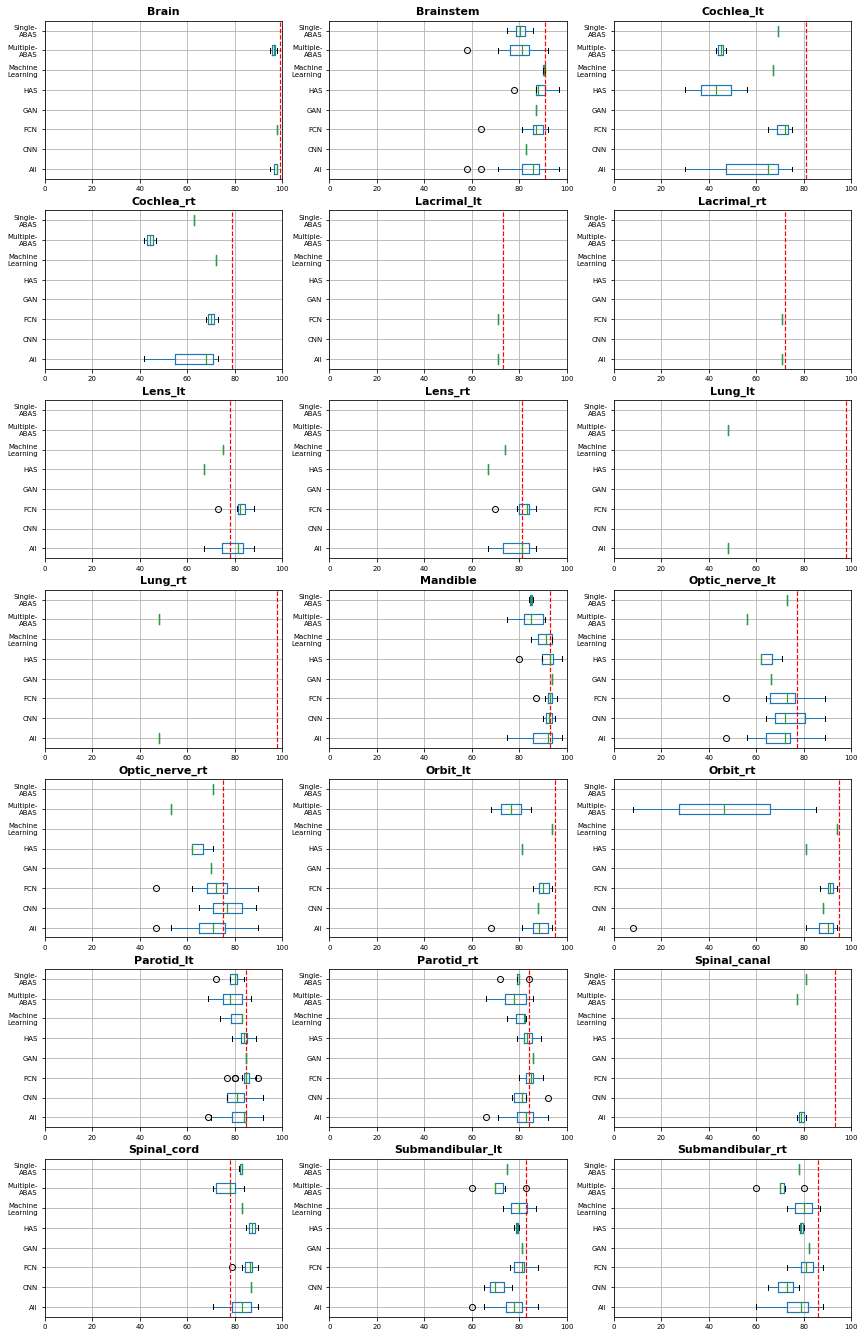}
    \caption{\textbf{Comparison of volumetric DSC performance or our model and previously published results}. The volumetric-DSC performance distribution is shown for each OAR. The performance distribution is shown for each method family and for all methods collectively. The blue boxes indicate the 1st and 3rd quartiles around the median (marked in green). The whiskers indicate most extreme, non-outlier data points. The red vertical lines indicate the performance of our model on the UCLH data.}
    \label{fig:perf_comparison}
\end{figure}

\end{document}

%% file: table_literature.tex
\setlength{\tabcolsep}{1pt}
\begin{tabular}{ll*{24}{c}}\toprule Study & \rotatebox{90}{Brain} & \rotatebox{90}{Brainstem} & \multicolumn{2}{c}{\rotatebox{90}{Cochlea}} & \multicolumn{2}{c}{\rotatebox{90}{Lacrimal}} & \multicolumn{2}{c}{\rotatebox{90}{Lens}} & \multicolumn{2}{c}{\rotatebox{90}{Lung}} & \rotatebox{90}{Mandible} &\multicolumn{2}{c}{\rotatebox{90}{Optic Nerve}} & \multicolumn{2}{c}{\rotatebox{90}{Orbit}} & \multicolumn{2}{c}{\rotatebox{90}{Parotid}} & \rotatebox{90}{Spinal Canal} &\rotatebox{90}{Spinal Cord} &\multicolumn{2}{c}{\rotatebox{90}{Submandibular}}\\\cmidrule(lr){4-5}\cmidrule(lr){6-7}\cmidrule(lr){8-9}\cmidrule(lr){10-11}\cmidrule(lr){13-14}\cmidrule(lr){15-16}\cmidrule(lr){17-18}\cmidrule(lr){21-22}&  &  & lt & rt & lt & rt & lt & rt & lt & rt & &lt & rt & lt & rt & lt & rt & &&lt & rt \\\midrule\\
Guo (2020) \cite{gou2020self}&&88&&&&&&&&&94&72&71&&&87&86&&&78&81\\
Liang (2020) \cite{liang2020multi}&&92&&&&&88&87&&&94&\multicolumn{2}{c}{74}&93&93&\multicolumn{2}{c}{88}&&90&\multicolumn{2}{c}{81}\\
Qiu (2020) \cite{qiu2020recurrent}&&&&&&&&&&&95&&&&&&&&&&\\
Sun (2020) \cite{sun2020attentionanatomy}&&86&&&&&&&&&94&&&90&90&84&81&&89&78&77\\
van Dijk (2020) \cite{van2020improving}&&83$^1$&&&&&&&&&95$^1$&&&&&84$^1$&83$^1$&&87$^1$&77$^1$&78$^1$\\
Wong (2020) \cite{wong2020comparing}&&83&&&&&&&&&&\multicolumn{2}{c}{47}&&&\multicolumn{2}{c}{80}&&79&\multicolumn{2}{c}{82}\\
Chan (2019) \cite{chan2019convolutional}&&89&&&&&&&&&91&&&&&85&86&&87&84&85\\
Gao (2019) \cite{gao2019focusnet}&&86&&&&&81&79&&&&64&62&88&91&77&80&&87&&\\
Jiang (2019) \cite{jiang2019local}&&88&&&&&&&&&93&&&&&85&86&&&79&77\\
Lei (2019) \cite{lei2019deepigeos}&&87&&&&&&&&&&\multicolumn{2}{c}{66}&&&\multicolumn{2}{c}{86}&&&&\\
Men (2019) \cite{men2019}&&90&&&&&&&&&92&&&&&86&86&&&&\\
Rhee (2019) \cite{rhee2019automatic}&98&86&65&68&&&73&70&&&87&89&90&89&90&83&83&&83&&\\
Sun (2019) \cite{sun2019accurate}&&&&&&&85&84&&&&80&82&94&94&&&&&&\\
Tang (2019) \cite{tang2019clinically}&&86&&&&&82&83&&&93&75&76&92&92&85&85&&86&81&83\\
Tappeiner (2019) \cite{tappeiner2019multi}&&82&&&&&&&&&91&64&63&&&80&81&&&&\\
Tong (2019) \cite{tong2019shape}&&87&&&&&&&&&94&66&70&&&85&86&&&81&82\\
van Rooij (2019) \cite{van2019deep}&&64&&&&&&&&&&&&&&83&83&&&82&81\\
Wang (2019) \cite{wang2019organ}&&88&&&&&&&&&93&74&74&&&86&85&&&76&73\\
Xue (2019) \cite{xue2019shape}&&90&&&&&&&&&96&86&84&&&89&89&&&86&85\\
Zhong (2019) \cite{zhong2019boosting}&&&&&&&&&&&&\multicolumn{2}{c}{89}&&&\multicolumn{2}{c}{92}&&&&\\
Hänsch (2018) \cite{Hansch2018-jo}&&&&&&&&&&&&&&&&\multicolumn{2}{c}{86}&&&&\\
Kodym (2018) \cite{kodym2018-ed}&&92&&&&&&&&&95&\multicolumn{2}{c}{80}&&&\multicolumn{2}{c}{90}&&&\multicolumn{2}{c}{88}\\
Liang (2018) \cite{Liang2018}&&90&&&&&83&84&&&91&66&72&&&85&85&&&&\\
Močnik (2018) \cite{Mocnik2018-fv}&&&&&&&&&&&&&&&&\multicolumn{2}{c}{77}&&&&\\
Nikolov (2018) \cite{nikolov2018deep}&99&88&65&75&69&70&81&80&99&99&96&76&77&95&95&85&85&95&88&85&85\\
Tong (2018) \cite{Tong2018}&&87&&&&&&&&&94&65&69&&&84&83&&&76&81\\
Ren (2018) \cite{Ren2018-vi}&&&&&&&&&&&&72&70&&&&&&&&\\
Willems (2018) \cite{willems2018-DeepVoxNet}&&92&75&73&&&&&&&96&&&&&86&90&&&79&88\\
Zhu (2018) \cite{Zhu2018-js}&&87&&&&&&&&&93&72&71&&&88&87&&&81&81\\
Ibragimov (2017) \cite{Ibragimov2017-is}&&&&&&&&&&&90&64&65&88&88&77&78&&87&70&73\\
Fritscher (2016) \cite{Fritscher2016-nd}&&&&&&&&&&&&&&&&\multicolumn{2}{c}{81}&&&\multicolumn{2}{c}{65}\\
\midrule
Radiographer (TCIA) &99.1&90.0&74.9&69.6&67.3&67.8&87.7&84.5&98.7&98.9&94.2&79.3&78.4&93.3&93.4&87.1&87.4&93.9&84.3&84.7&77.5\\
(28 scans)&$\pm$0.2&$\pm$2.5&$\pm$10.9&$\pm$23.1&$\pm$10.4&$\pm$11.0&$\pm$8.0&$\pm$14.7&$\pm$0.7&$\pm$0.5&$\pm$2.2&$\pm$4.9&$\pm$6.2&$\pm$2.1&$\pm$1.9&$\pm$3.4&$\pm$3.1&$\pm$1.8&$\pm$4.6&$\pm$18.3&$\pm$28.5\\
\\
Our model (TCIA) &98.8&85.1&80.5&81.0&64.4&63.8&81.6&75.7&98.7&98.8&92.9&77.9&76.3&92.6&93.1&84.1&84.6&91.7&80.3&81.8&77.8\\
(28 scans)&$\pm$1.1&$\pm$7.1&$\pm$8.8&$\pm$7.2&$\pm$11.9&$\pm$9.0&$\pm$16.6&$\pm$24.5&$\pm$0.6&$\pm$0.7&$\pm$3.5&$\pm$5.0&$\pm$5.8&$\pm$2.0&$\pm$1.8&$\pm$5.8&$\pm$4.2&$\pm$1.6&$\pm$7.6&$\pm$8.7&$\pm$18.1\\
\\
\midrule
Radiographer (UCLH) &99.2&90.1&77.9&80.3&74.1&71.8&82.7&83.9&98.6&98.6&95.8&80.3&79.4&93.9&94.2&88.1&87.5&93.1&81.6&87.5&86.8\\
(21 scans)&$\pm$0.2&$\pm$2.4&$\pm$14.0&$\pm$10.1&$\pm$7.0&$\pm$7.8&$\pm$22.6&$\pm$23.8&$\pm$0.9&$\pm$1.3&$\pm$1.2&$\pm$5.2&$\pm$7.4&$\pm$1.4&$\pm$0.9&$\pm$2.8&$\pm$3.4&$\pm$2.0&$\pm$6.0&$\pm$4.0&$\pm$4.0\\
\\
Our model (UCLH) &99&91&81&79&73&72&78&81&98&98&93&77&75&95&95&85&84&93&78&83&86\\
(21 scans)&$\pm$0.2&$\pm$2.2&$\pm$8.2&$\pm$5.7&$\pm$5.6&$\pm$5.8&$\pm$25.0&$\pm$25.8&$\pm$1.3&$\pm$2.2&$\pm$2.0&$\pm$4.8&$\pm$7.0&$\pm$1.3&$\pm$1.0&$\pm$3.8&$\pm$4.5&$\pm$1.4&$\pm$8.9&$\pm$8.4&$\pm$4.9\\
\\
Oncologist (UCLH) &99.0$^3$&91.9$^3$&68.5&75.8&63.3&61.6&86.2&87.6&98.4$^3$&98.6$^3$&95.4$^3$&77.1&76.0&94.8$^3$&94.8$^3$&90.1$^3$&90.7$^3$&94.9$^3$&87.7$^3$&91.1$^3$&90.1$^3$\\
(8 - 75 scans)$^2$&&&$\pm$14.8&$\pm$8.5&$\pm$13.1&$\pm$14.3&$\pm$10.1&$\pm$9.9&&&&$\pm$6.3&$\pm$7.1&&&&&&&&\\
\midrule
\\
Our model (PDDCA) &&84.2&&&&&&&&&93.8&71.6&69.1&&&88.1&86.6&&&76.5&79.2\\
(15 scans)&&$\pm$5.2&&&&&&&&&$\pm$1.9&$\pm$6.2&$\pm$5.9&&&$\pm$2.0&$\pm$3.5&&&$\pm$9.1&$\pm$6.5\\
\bottomrule
\end{tabular}

%% file: table_organs.tex
\begin{tabularx}{0.9\textwidth}{l  c  p{8cm}}
\toprule
OAR              & 
\parbox{3cm}{\centering{Total number of labelled slices included}} &
Anatomical Landmarks and Definition\\
\midrule
Brain            & 11476                                     & Sits inside the cranium and includes all brain vessels excluding the brainstem and optic chiasm.\\

\rule{0pt}{3ex}%
Brainstem        & 34794                                     & The posterior aspect of the brain including the midbrain, pons and medulla oblongata. Extending inferior from the lateral ventricles to the tip of the dens at C2. It is structurally continuous with the spinal cord.\\

\rule{0pt}{3ex}%
Cochlea-Lt       & 4526                                     & \multirow{2}{\linewidth}{Embedded in the temporal bone and lateral to the internal auditory meatus.}\\

Cochlea-Rt       & 4754                                     & \\

\rule{0pt}{3ex}%
Lacrimal-Lt      & 17186                                    & \multirow{2}{\linewidth}{Concave shaped gland located at the superolateral aspect of the orbit.}\\

Lacrimal-Rt      & 17788                                    & \\

\rule{0pt}{3ex}%
Lens-Lt          & 3006                                     & \multirow{2}{\linewidth}{An oval structure that sits within the anterior segment of the orbit. Can be variable in position but never sitting posterior beyond the level of the outer canthus.} \\

Lens-Rt          & 3354
 & \\
 && \\

\rule{0pt}{3ex}%
Lung-Lt          & 8340                                     & \multirow{2}{\linewidth}{\parbox{\linewidth}{Encompassed by the thoracic cavity adjacent to the lateral aspect of the mediastinum, extending from the 1st rib to the diaphragm excluding the carina.}}\\

Lung-Rt          & 9158                                     & \\
 && \\
 
\rule{0pt}{3ex}%
Mandible         & 25074                                     & The entire mandible bone including the temporomandibular joint, ramus and body, excluding the teeth. The mandible joins to the inferior aspect of the temporal bone and forms the entire lower jaw.\\

\rule{0pt}{3ex}%
Optic-Nerve-Lt   & 3458                                     & \multirow{2}{\linewidth}{A 2-5mm thick nerve that runs from the posterior aspect of the eye, through the optic canal and ends at the lateral aspect of the optic chiasm.}\\

Optic-Nerve-Rt   & 3012                                     &\\
 && \\

\rule{0pt}{3ex}%
Orbit-Lt         & 8538                                     & \multirow{2}{\linewidth}{Spherical organ sitting within the orbital cavity. Includes the vitreous humor, retina, cornea and lens with the optic nerve attached posteriorly.}\\

Orbit-Rt         & 8242                                     &\\
 && \\

\rule{0pt}{3ex}%
Parotid-Lt       & 8984                                     & \multirow{2}{\linewidth}{Multi lobed salivary gland wrapped around the mandibular ramus. Extends medially to styloid process and parapharyngeal space. Laterally extending to subcutaneous fat. Posteriorly extending to sternocleidomastoid muscle. Anterior extending to posterior border of mandible bone and masseter muscle. In cases where retromandibular vein is encapsulated by parotid this is included in the segmentation.} \\

Parotid-Rt       & 11752                                     &\\
 && \\
 && \\
 && \\
 && \\
 && \\

\rule{0pt}{3ex}%
Spinal-Canal     & 37000                                    & Hollow cavity that runs through the foramen of the vertebrae, extending from the base of skull to the end of the sacrum.\\

\rule{0pt}{3ex}%
Spinal-Cord      & 37096                                    & Sits inside the Spinal Canal and extends from the level of the foramen magnum to the bottom of L2.\\

\rule{0pt}{3ex}%
Submandibular-Lt & 10652                                     & \multirow{5}{\linewidth}{Sits within the submandibular portion of the anterior triangle of the neck, making up the floor of the mouth and extending both superior and inferior to the posterior aspect of the mandible and is limited laterally by the mandible and medially by the hypoglossal muscle.}\\

Submandibular-Rt & 10716                                     & \\
 && \\
 && \\
 && \\

\bottomrule
\end{tabularx}

%% file: table_surface_dice_TCIA.tex
\setlength{\tabcolsep}{1pt}
\begin{tabular}{ll*{28}{r}@{\hspace{2em}}rr}\toprule
&&\multicolumn{28}{c}{TCIA test set patient ID}\\
\cmidrule(lr){3-30}
Organ&M/H&\rotatebox{90}{0522c 0017}&\rotatebox{90}{0522c 0057}&\rotatebox{90}{0522c 0161}&\rotatebox{90}{0522c 0226}&\rotatebox{90}{0522c 0248}&\rotatebox{90}{0522c 0251}&\rotatebox{90}{0522c 0331}&\rotatebox{90}{0522c 0416}&\rotatebox{90}{0522c 0419}&\rotatebox{90}{0522c 0427}&\rotatebox{90}{0522c 0457}&\rotatebox{90}{0522c 0479}&\rotatebox{90}{0522c 0629}&\rotatebox{90}{0522c 0659}&\rotatebox{90}{0522c 0667}&\rotatebox{90}{0522c 0669}&\rotatebox{90}{0522c 0708}&\rotatebox{90}{0522c 0768}&\rotatebox{90}{0522c 0770}&\rotatebox{90}{0522c 0773}&\rotatebox{90}{0522c 0845}&\rotatebox{90}{TCGA-CV-7236}&\rotatebox{90}{TCGA-CV-7243}&\rotatebox{90}{TCGA-CV-7245}&\rotatebox{90}{TCGA-CV-A6JO}&\rotatebox{90}{TCGA-CV-A6JY}&\rotatebox{90}{TCGA-CV-A6K0}&\rotatebox{90}{TCGA-CV-A6K1}&mean, stddev&diff.\\
\midrule
\raisebox{-4pt}[-4pt][-4pt]{\textbf{Brain}}&(M)&95&94&96&94&96&94&96&98&94&\cellcolor{magenta!50}71&98&95&97&94&96&97&97&98&91&94&97&97&97&97&97&96&96&94&94.9$\pm$4.8\\
\raisebox{-4pt}[-4pt][-4pt]{(84054.9 mm$^2$)}&(H)&97&96&97&95&93&96&96&97&95&\cellcolor{magenta!50}96&98&96&97&97&97&96&96&98&96&97&97&98&96&96&97&96&95&95&96.3$\pm$1.0&\raisebox{4pt}[-4pt]{-1.3}\\
\\
\raisebox{-4pt}[-4pt][-4pt]{\textbf{Brainstem}}&(M)&\cellcolor{magenta!50}84&\cellcolor{magenta!50}72&92&\cellcolor{magenta!50}79&\cellcolor{magenta!25}86&\cellcolor{magenta!50}65&92&94&\cellcolor{magenta!50}87&\cellcolor{magenta!50}62&99&93&97&99&99&100&99&98&\cellcolor{magenta!50}72&\cellcolor{magenta!50}82&95&97&98&\cellcolor{magenta!50}87&98&96&94&94&\cellcolor{magenta!25}89.5$\pm$10.5\\
\bottomrule
\end{tabular}

%% file: table_volumetric_dice_TCIA.tex
\setlength{\tabcolsep}{2.5pt}
\begin{tabular}{ll*{28}{r}@{\hspace{2em}}rr}\toprule
&&\multicolumn{28}{c}{TCIA test set patient ID}\\
\cmidrule(lr){3-30}
Organ&M/H&\rotatebox{90}{0522c 0017}&\rotatebox{90}{0522c 0057}&\rotatebox{90}{0522c 0161}&\rotatebox{90}{0522c 0226}&\rotatebox{90}{0522c 0248}&\rotatebox{90}{0522c 0251}&\rotatebox{90}{0522c 0331}&\rotatebox{90}{0522c 0416}&\rotatebox{90}{0522c 0419}&\rotatebox{90}{0522c 0427}&\rotatebox{90}{0522c 0457}&\rotatebox{90}{0522c 0479}&\rotatebox{90}{0522c 0629}&\rotatebox{90}{0522c 0659}&\rotatebox{90}{0522c 0667}&\rotatebox{90}{0522c 0669}&\rotatebox{90}{0522c 0708}&\rotatebox{90}{0522c 0768}&\rotatebox{90}{0522c 0770}&\rotatebox{90}{0522c 0773}&\rotatebox{90}{0522c 0845}&\rotatebox{90}{TCGA-CV-7236}&\rotatebox{90}{TCGA-CV-7243}&\rotatebox{90}{TCGA-CV-7245}&\rotatebox{90}{TCGA-CV-A6JO}&\rotatebox{90}{TCGA-CV-A6JY}&\rotatebox{90}{TCGA-CV-A6K0}&\rotatebox{90}{TCGA-CV-A6K1}&mean, stddev&diff.\\
\midrule
\raisebox{-4pt}[-4pt][-4pt]{\textbf{Brain}}&(M)&\cellcolor{magenta!15}99&\cellcolor{magenta!15}99&\cellcolor{magenta!15}99&\cellcolor{magenta!15}99&\cellcolor{green!15}99&\cellcolor{magenta!15}99&\cellcolor{green!15}99&\cellcolor{green!15}99&\cellcolor{green!15}99&\cellcolor{magenta!15}93&\cellcolor{magenta!15}99&\cellcolor{green!15}99&\cellcolor{magenta!15}99&\cellcolor{magenta!15}99&\cellcolor{magenta!15}99&\cellcolor{magenta!15}99&\cellcolor{magenta!15}99&\cellcolor{magenta!15}99&\cellcolor{magenta!15}99&\cellcolor{magenta!15}99&\cellcolor{green!15}99&\cellcolor{green!15}99&\cellcolor{green!15}99&\cellcolor{magenta!15}99&\cellcolor{magenta!15}99&\cellcolor{green!15}99&\cellcolor{green!15}99&\cellcolor{magenta!15}99&\cellcolor{magenta!15}99$\pm$ 1\\
\bottomrule
\end{tabular}

%% file: table_surface_dice_pddca.tex
\setlength{\tabcolsep}{3pt}
\begin{tabular}{l*{15}{r}@{\hspace{2em}}rr}\toprule
&\multicolumn{15}{c}{PDDCA test set patient ID}\\
\cmidrule(lr){2-16}
&\multicolumn{10}{c}{off-site test set}&\multicolumn{5}{c}{on-site test set}\\
\cmidrule(lr){2-11}\cmidrule(lr){12-16}
Organ&\rotatebox{67.5}{0522c 0555}&\rotatebox{67.5}{0522c 0576}&\rotatebox{67.5}{0522c 0598}&\rotatebox{67.5}{0522c 0659}&\rotatebox{67.5}{0522c 0661}&\rotatebox{67.5}{0522c 0667}&\rotatebox{67.5}{0522c 0669}&\rotatebox{67.5}{0522c 0708}&\rotatebox{67.5}{0522c 0727 c}&\rotatebox{67.5}{0522c 0746}&\rotatebox{67.5}{0522c 0788}&\rotatebox{67.5}{0522c 0806}&\rotatebox{67.5}{0522c 0845}&\rotatebox{67.5}{0522c 0857}&\rotatebox{67.5}{0522c 0878}&mean, stddev\\
\midrule
\textbf{Brainstem}&84.4&85.4&87.4&89.9&\cellcolor{green!50}98.4&\cellcolor{magenta!25}79.4&\cellcolor{green!50}98.9&\cellcolor{green!50}99.9&\cellcolor{green!50}98.1&\cellcolor{magenta!50}72.3&\cellcolor{green!50}98.2&89.7&\cellcolor{green!25}91.2&\cellcolor{green!50}95.0&\cellcolor{magenta!50}71.5&89.3$\pm$9.0\\
(5042.8 mm$^2$)\\
\\
\textbf{Mandible}&\cellcolor{green!50}96.1&\cellcolor{green!50}98.5&\cellcolor{green!50}97.9&\cellcolor{green!50}97.5&\cellcolor{green!50}96.6&\cellcolor{green!50}98.4&\cellcolor{green!50}96.4&\cellcolor{green!50}99.8&\cellcolor{green!50}98.6&\cellcolor{green!50}98.3&\cellcolor{green!50}97.0&\cellcolor{green!25}93.7&\cellcolor{green!50}97.6&\cellcolor{green!50}96.2&\cellcolor{green!25}94.1&\cellcolor{green!50}97.1$\pm$1.6\\
(17215.2 mm$^2$)\\
\bottomrule
\end{tabular}

%% file: table_volumetric_dice_pddca.tex
\setlength{\tabcolsep}{3pt}
\begin{tabular}{l*{15}{r}@{\hspace{2em}}rr}\toprule
&\multicolumn{15}{c}{PDDCA test set patient ID}\\
\cmidrule(lr){2-16}
&\multicolumn{10}{c}{off-site test set}&\multicolumn{5}{c}{on-site test set}\\
\cmidrule(lr){2-11}\cmidrule(lr){12-16}
Organ&\rotatebox{67.5}{0522c 0555}&\rotatebox{67.5}{0522c 0576}&\rotatebox{67.5}{0522c 0598}&\rotatebox{67.5}{0522c 0659}&\rotatebox{67.5}{0522c 0661}&\rotatebox{67.5}{0522c 0667}&\rotatebox{67.5}{0522c 0669}&\rotatebox{67.5}{0522c 0708}&\rotatebox{67.5}{0522c 0727 c}&\rotatebox{67.5}{0522c 0746}&\rotatebox{67.5}{0522c 0788}&\rotatebox{67.5}{0522c 0806}&\rotatebox{67.5}{0522c 0845}&\rotatebox{67.5}{0522c 0857}&\rotatebox{67.5}{0522c 0878}&mean, stddev\\
\midrule
\textbf{Brainstem}&82.0&82.7&83.0&84.9&88.8&\cellcolor{magenta!25}76.4&88.5&\cellcolor{green!25}92.8&86.7&\cellcolor{magenta!25}76.3&89.9&85.3&86.5&85.5&\cellcolor{magenta!50}73.7&84.2$\pm$5.2\\
(19778.8 mm$^3$)\\
\\
\textbf{Mandible}&\cellcolor{green!25}94.2&\cellcolor{green!25}92.1&\cellcolor{green!50}95.8&\cellcolor{green!25}94.9&\cellcolor{green!25}90.4&\cellcolor{green!25}94.9&\cellcolor{green!25}94.6&\cellcolor{green!50}96.3&\cellcolor{green!50}96.0&\cellcolor{green!50}95.4&\cellcolor{green!25}92.8&\cellcolor{green!25}90.8&\cellcolor{green!25}92.5&\cellcolor{green!25}94.7&\cellcolor{green!25}91.6&\cellcolor{green!25}93.8$\pm$1.9\\
(44477.1 mm$^3$)\\
\bottomrule
\end{tabular}

%% file: table_surface_dice_UCLH.tex
\setlength{\tabcolsep}{1pt}
\begin{tabular}{ll*{21}{r}@{\hspace{2em}}rr}\toprule
&&\multicolumn{21}{c}{UCLH test set patient ID}\\
\cmidrule(lr){3-23}
Organ&M/H&\rotatebox{90}{UCLH-01}&\rotatebox{90}{UCLH-02}&\rotatebox{90}{UCLH-03}&\rotatebox{90}{UCLH-04}&\rotatebox{90}{UCLH-05}&\rotatebox{90}{UCLH-06}&\rotatebox{90}{UCLH-07}&\rotatebox{90}{UCLH-08}&\rotatebox{90}{UCLH-09}&\rotatebox{90}{UCLH-10}&\rotatebox{90}{UCLH-11}&\rotatebox{90}{UCLH-12}&\rotatebox{90}{UCLH-13}&\rotatebox{90}{UCLH-14}&\rotatebox{90}{UCLH-15}&\rotatebox{90}{UCLH-16}&\rotatebox{90}{UCLH-17}&\rotatebox{90}{UCLH-18}&\rotatebox{90}{UCLH-19}&\rotatebox{90}{UCLH-20}&\rotatebox{90}{UCLH-21}&mean, stddev&diff.\\
\midrule
\raisebox{-4pt}[-4pt][-4pt]{\textbf{Brain}}&(M)&98&97&94&95&94&96&96&94&94&96&95&95&94&96&94&97&97&95&93&97&96&95.5$\pm$1.3\\
\raisebox{-4pt}[-4pt][-4pt]{(81738.4 mm$^2$)}&(H)&97&99&94&96&98&96&97&96&96&96&94&97&95&95&95&99&98&97&96&97&98&96.4$\pm$1.3&\raisebox{4pt}[-4pt]{-0.9}\\
\\
\raisebox{-4pt}[-4pt][-4pt]{\textbf{Brainstem}}&(M)&100&99&\cellcolor{magenta!50}84&97&99&98&99&99&97&99&98&98&99&\cellcolor{green!25}98&98&97&99&99&98&97&100&97.7$\pm$3.2\\
\raisebox{-4pt}[-4pt][-4pt]{(6555.3 mm$^2$)}&(H)&98&99&\cellcolor{magenta!50}95&95&98&98&96&97&97&97&95&98&95&\cellcolor{green!25}91&97&99&99&99&100&95&99&97.0$\pm$2.1&\raisebox{4pt}[-4pt]{0.7}\\
\\
\bottomrule
\end{tabular}

%% file: table_volumetric_dice_UCLH.tex
\setlength{\tabcolsep}{2.5pt}
\begin{tabular}{ll*{21}{r}@{\hspace{2em}}rr}\toprule
&&\multicolumn{21}{c}{UCLH test set patient ID}\\
\cmidrule(lr){3-23}
Organ&M/H&\rotatebox{90}{UCLH-01}&\rotatebox{90}{UCLH-02}&\rotatebox{90}{UCLH-03}&\rotatebox{90}{UCLH-04}&\rotatebox{90}{UCLH-05}&\rotatebox{90}{UCLH-06}&\rotatebox{90}{UCLH-07}&\rotatebox{90}{UCLH-08}&\rotatebox{90}{UCLH-09}&\rotatebox{90}{UCLH-10}&\rotatebox{90}{UCLH-11}&\rotatebox{90}{UCLH-12}&\rotatebox{90}{UCLH-13}&\rotatebox{90}{UCLH-14}&\rotatebox{90}{UCLH-15}&\rotatebox{90}{UCLH-16}&\rotatebox{90}{UCLH-17}&\rotatebox{90}{UCLH-18}&\rotatebox{90}{UCLH-19}&\rotatebox{90}{UCLH-20}&\rotatebox{90}{UCLH-21}&mean, stddev&diff.\\
\midrule
\raisebox{-4pt}[-4pt][-4pt]{\textbf{Brain}}&(M)&\cellcolor{magenta!15}99&\cellcolor{magenta!15}99&\cellcolor{green!15}99&\cellcolor{magenta!15}99&\cellcolor{magenta!15}99&\cellcolor{magenta!15}99&\cellcolor{magenta!15}99&\cellcolor{green!15}99&\cellcolor{magenta!15}99&\cellcolor{magenta!15}99&\cellcolor{green!15}99&\cellcolor{magenta!15}99&\cellcolor{green!15}99&\cellcolor{green!15}99&\cellcolor{green!15}99&\cellcolor{magenta!15}99&\cellcolor{magenta!15}99&\cellcolor{magenta!15}99&\cellcolor{magenta!15}99&\cellcolor{magenta!15}99&\cellcolor{magenta!15}99&\cellcolor{magenta!15}99$\pm$ 0\\
\raisebox{-4pt}[-4pt][-4pt]{(1316891.7 mm$^3$)}&(H)&\cellcolor{magenta!15}99&\cellcolor{magenta!15}99&\cellcolor{green!15}99&\cellcolor{magenta!15}99&\cellcolor{magenta!15}99&\cellcolor{magenta!15}99&\cellcolor{magenta!15}99&\cellcolor{green!15}99&\cellcolor{magenta!15}99&\cellcolor{magenta!15}99&\cellcolor{green!15}99&\cellcolor{magenta!15}99&\cellcolor{green!15}99&\cellcolor{green!15}99&\cellcolor{green!15}99&\cellcolor{magenta!15}99&\cellcolor{magenta!15}99&\cellcolor{magenta!15}99&\cellcolor{magenta!15}99&\cellcolor{magenta!15}99&\cellcolor{magenta!15}99&\cellcolor{magenta!15}99.2$\pm$0.17&\raisebox{4pt}[-4pt]{-0.1}\\
\\
\bottomrule
\end{tabular}

%% file: table_num_scans_UCLH.tex
\setlength{\tabcolsep}{3pt}
  \begin{tabular}{ll*{24}{c}}\toprule
  && 
  Brain & 
  Brainstem & 
  \multicolumn{2}{c}{Cochlea} & 
  \multicolumn{2}{c}{Lacrimal} & 
  \multicolumn{2}{c}{Lens} & 
  \multicolumn{2}{c}{Lung} & 
  Mandible &
  \multicolumn{2}{c}{Optic Nerve} & 
  \multicolumn{2}{c}{Orbit} & 
  \multicolumn{2}{c}{Parotid} & 
  Spinal-Canal &
  Spinal-Cord &
  \multicolumn{2}{c}{Submandibular}\\
  \cmidrule(lr){5-6}
  \cmidrule(lr){7-8}
  \cmidrule(lr){9-10}
  \cmidrule(lr){11-12}
  \cmidrule(lr){14-15}
  \cmidrule(lr){16-17}
  \cmidrule(lr){18-19}
  \cmidrule(lr){22-23}
  &  &  &  & 
  lt & rt & 
  lt & rt & 
  lt & rt & 
  lt & rt & 
  &
  lt & rt & 
  lt & rt & 
  lt & rt & 
  &
  &
  lt & rt \\
  \midrule
  Number of scans&&75&45&8&8&75&73&75&73&71&72&74&17&15&19&16&33&32&23&24&64&65\\
Dense segmentation&&&&$\checkmark$&$\checkmark$&$\checkmark$&$\checkmark$&$\checkmark$&$\checkmark$&&&&$\checkmark$&$\checkmark$&&&&&&&&\\
Number of labelled slices& axial&309&225&&&&&&&265&275&300&&&95&75&165&160&345&350&250&260\\
& coronal&374&225&&&&&&&355&360&375&&&95&80&165&160&0&0&320&325\\
& sagittal&374&225&&&&&&&355&360&375&&&95&80&165&160&0&0&320&325\\
\bottomrule
\end{tabular}

%% file: table_literature_full.tex
\setlength{\tabcolsep}{1pt}\hspace{-8mm}\begin{tabular}{ll*{24}{c}}\toprule Study & Method & \rotatebox{90}{Brain} & \rotatebox{90}{Brainstem} & \multicolumn{2}{c}{\rotatebox{90}{Cochlea}} & \multicolumn{2}{c}{\rotatebox{90}{Lacrimal}} & \multicolumn{2}{c}{\rotatebox{90}{Lens}} & \multicolumn{2}{c}{\rotatebox{90}{Lung}} & \rotatebox{90}{Mandible} &\multicolumn{2}{c}{\rotatebox{90}{Optic Nerve}} & \multicolumn{2}{c}{\rotatebox{90}{Orbit}} & \multicolumn{2}{c}{\rotatebox{90}{Parotid}} & \rotatebox{90}{Spinal Canal} &\rotatebox{90}{Spinal Cord} &\multicolumn{2}{c}{\rotatebox{90}{Submandibular}}\\\cmidrule(lr){5-6}\cmidrule(lr){7-8}\cmidrule(lr){9-10}\cmidrule(lr){11-12}\cmidrule(lr){14-15}\cmidrule(lr){16-17}\cmidrule(lr){18-19}\cmidrule(lr){22-23}&  &  &  & lt & rt & lt & rt & lt & rt & lt & rt & &lt & rt & lt & rt & lt & rt & &&lt & rt \\\midrule\\
van Dijk (2020) \cite{van2020improving}&CNN&&83$^2$&&&&&&&&&95$^2$&&&&&84$^2$&83$^2$&&87$^2$&77$^2$&78$^2$\\
Zhong (2019) \cite{zhong2019boosting}&CNN&&&&&&&&&&&&\multicolumn{2}{c}{89}&&&\multicolumn{2}{c}{92}&&&&\\
Močnik (2018) \cite{Mocnik2018-fv}&CNN&&&&&&&&&&&&&&&&\multicolumn{2}{c}{77}&&&&\\
Ren (2018) \cite{Ren2018-vi}&CNN&&&&&&&&&&&&72&70&&&&&&&&\\
Ibragimov (2017) \cite{Ibragimov2017-is}&CNN&&&&&&&&&&&90&64&65&88&88&77&78&&87&70&73\\
Fritscher (2016) \cite{Fritscher2016-nd}&CNN&&&&&&&&&&&&&&&&\multicolumn{2}{c}{81}&&&\multicolumn{2}{c}{65}\\
Guo (2020) \cite{gou2020self}&FCN&&88&&&&&&&&&94&72&71&&&87&86&&&78&81\\
Qiu (2020) \cite{qiu2020recurrent}&FCN&&&&&&&&&&&95&&&&&&&&&&\\
Sun (2020) \cite{sun2020attentionanatomy}&FCN&&86&&&&&&&&&94&&&90&90&84&81&&89&78&77\\
Wong (2020) \cite{wong2020comparing}&FCN&&83&&&&&&&&&&\multicolumn{2}{c}{47}&&&\multicolumn{2}{c}{80}&&79&\multicolumn{2}{c}{82}\\
Liang (2020) \cite{liang2020multi}&FCN&&92&&&&&88&87&&&94&\multicolumn{2}{c}{74}&93&93&\multicolumn{2}{c}{88}&&90&\multicolumn{2}{c}{81}\\
Xue (2019) \cite{xue2019shape}&FCN&&90&&&&&&&&&96&86&84&&&89&89&&&86&85\\
Chan (2019) \cite{chan2019convolutional}&FCN&&89&&&&&&&&&91&&&&&85&86&&87&84&85\\
Gao (2019) \cite{gao2019focusnet}&FCN&&86&&&&&81&79&&&&64&62&88&91&77&80&&87&&\\
Lei (2019) \cite{lei2019deepigeos}&FCN&&87&&&&&&&&&&\multicolumn{2}{c}{66}&&&\multicolumn{2}{c}{86}&&&&\\
Sun (2019) \cite{sun2019accurate}&FCN&&&&&&&85&84&&&&80&82&94&94&&&&&&\\
Jiang (2019) \cite{jiang2019local}&FCN&&88&&&&&&&&&93&&&&&85&86&&&79&77\\
van Rooij (2019) \cite{van2019deep}&FCN&&64&&&&&&&&&&&&&&83&83&&&82&81\\
Tang (2019) \cite{tang2019clinically}&FCN&&86&&&&&82&83&&&93&75&76&92&92&85&85&&86&81&83\\
Rhee (2019) \cite{rhee2019automatic}&FCN&98&86&65&68&&&73&70&&&87&89&90&89&90&83&83&&83&&\\
Tappeiner (2019) \cite{tappeiner2019multi}&FCN&&82&&&&&&&&&91&64&63&&&80&81&&&&\\
Men (2019) \cite{men2019}&FCN&&90&&&&&&&&&92&&&&&86&86&&&&\\
Wang (2019) \cite{wang2019organ}&FCN&&88&&&&&&&&&93&74&74&&&86&85&&&76&73\\
Nikolov (2018) \cite{nikolov2018deep}&FCN&99&88&65&75&69&70&81&80&99&99&96&76&77&95&95&85&85&95&88&85&85\\
Kodym (2018) \cite{kodym2018-ed}&FCN&&92&&&&&&&&&95&\multicolumn{2}{c}{80}&&&\multicolumn{2}{c}{90}&&&\multicolumn{2}{c}{88}\\
Tong (2018) \cite{Tong2018}&FCN&&87&&&&&&&&&94&65&69&&&84&83&&&76&81\\
Zhu (2018) \cite{Zhu2018-js}&FCN&&87&&&&&&&&&93&72&71&&&88&87&&&81&81\\
Willems (2018) \cite{willems2018-DeepVoxNet}&FCN&&92&75&73&&&&&&&96&&&&&86&90&&&79&88\\
Hänsch (2018) \cite{Hansch2018-jo}&FCN&&&&&&&&&&&&&&&&\multicolumn{2}{c}{86}&&&&\\
Liang (2018) \cite{Liang2018}&FCN&&90&&&&&83&84&&&91&66&72&&&85&85&&&&\\
Tong (2019) \cite{tong2019shape}&GAN&&87&&&&&&&&&94&66&70&&&85&86&&&81&82\\
Gacha (2018) \cite{Gacha2018}&HAS&&&&&&&&&&&80&&&&&&&&&&\\
Raudashl (2017) \cite{Raudaschl2017-mn}&HAS&&88&&&&&&&&&93&\multicolumn{2}{c}{62}&&&\multicolumn{2}{c}{84}&&&\multicolumn{2}{c}{78}\\
Fritcher (2014) \cite{fritscher2014automatic}&HAS&&87$^{2,3}$&&&&&&&&&&&&&&84$^{2,3}$&83$^{2,3}$&&&&\\
Walker (2014) \cite{Walker2014-dd}&HAS&&97&\multicolumn{2}{c}{56}&&&&&&&98&\multicolumn{2}{c}{71}&&&\multicolumn{2}{c}{89}&&90&\multicolumn{2}{c}{73}\\
Thomson (2014) \cite{Thomson2014-pr}&HAS&&&\multicolumn{2}{c}{30$^{2,3}$}&&&&&&&&&&&&\multicolumn{2}{c}{79$^3$}&&&\multicolumn{2}{c}{80$^3$}\\
Fortunati (2013) \cite{Fortunati2013-ay}&HAS&&78&&&&&\multicolumn{2}{c}{67}&&&&\multicolumn{2}{c}{62}&\multicolumn{2}{c}{81}&&&&85&&\\
Qazi (2011) \cite{Qazi2011-ta}&HAS&&91&&&&&&&&&93&&&&&&&&&&\\
Wu (2019) \cite{WU2019}&Machine learning&&&&&&&&&&&89&&&&&74&75&&&73&73\\
Tam (2018) \cite{Tam2018-rr}&Machine learning&&91&67&72&&&75&74&&&85&&&94&94&83&82&&83&87&87\\
Wang (2017) \cite{Wang2018-eq}&Machine learning&&90&&&&&&&&&94&&&&&82&83&&&&\\
Torosdagli (2017) \cite{torosdagli2017robust}&Machine learning&&&&&&&&&&&94&&&&&&&&&&\\
Wang (2019) \cite{wang2019head}&Multi-ABAS&&84&&&&&&&&&75&56&53&&&75&74&&&74&72\\
Ayyalusamy (2019) \cite{ayyalusamy2019auto}&Multi-ABAS&&83$^2$&&&&&&&&&85$^2$&&&&&\multicolumn{2}{c}{81$^2$}&&84$^2$&&\\
Haq (2019) \cite{haq2019dynamic}&Multi-ABAS&&76&&&&&&&&&85&&&&&76&76&&84&60&60\\
McCarroll (2018) \cite{mccarroll2018retrospective}&Multi-ABAS&98&81&\multicolumn{2}{c}{47}&&&&&\multicolumn{2}{c}{48}&84&&&\multicolumn{2}{c}{85}&\multicolumn{2}{c}{78}&&71&&\\
Liu (2016) \cite{liu2016evaluation}&Multi-ABAS&&92&&&&&&&&&90&&&&&87&85&&80&83&80\\
Hoang Duc (2015) \cite{hoang2015validation}&Multi-ABAS&&82$^{2,3}$&&&&&&&&&&&&\multicolumn{2}{c}{68$^{2,3}$}&70$^{2,3}$&71$^{2,3}$&&79$^{2,3}$&&\\
Tao (2015) \cite{tao2015multi}&Multi-ABAS&&86&43&42&&&&&&&&&&&&&&77&&&\\
Wachinger (2015) \cite{wachinger2015contour}&Multi-ABAS&&&&&&&&&&&&&&&&78$^{2,3}$&77$^{2,3}$&&&&\\
Zhu (2013) \cite{zhu2013multi}&Multi-ABAS&95$^2$&72$^2$&&&&&&&&&90$^2$&&&&&\multicolumn{2}{c}{72$^2$}&&72$^2$&\multicolumn{2}{c}{70$^2$}\\
Teguh (2011) \cite{Teguh2011-dq}&Multi-ABAS&&78$^1$&&&&&&&&&&&&&&\multicolumn{2}{c}{79}&&78$^1$&\multicolumn{2}{c}{70}\\
Han (2010) \cite{han2010automatic}&Multi-ABAS&&&&&&&&&&&&&&&&85&86&&&&\\
Sims (2009) \cite{Sims2009-ou}&Multi-ABAS&&77&&&&&&&&&82&&&&&84&86&&&&\\
Sims (2009) \cite{sims2009pre}&Multi-ABAS&&58&&&&&&&&&78&&&&&69&66&&&&\\
Han (2008) \cite{han2008atlas}&Multi-ABAS&&84$^{2,3}$&&&&&&&&&91$^{2,3}$&&&&&\multicolumn{2}{c}{83$^{2,3}$}&&75$^{2,3}$&\multicolumn{2}{c}{70$^{2,3}$}\\
Hoogeman (2008) \cite{Hoogeman2008-dw}&Multi-ABAS&&71$^1$&&&&&&&&&&&&&&&&&71$^1$&&\\
Huang (2019) \cite{huang2019atlas}&Single-ABAS&&&&&&&&&&&84&73&71&&&84&84&&82&75&78\\
Daisne (2013) \cite{Daisne2013-uk}&Single-ABAS&&75$^2$&&&&&&&&&&&&&&\multicolumn{2}{c}{72$^2$}&&&&\\
Hardcastle (2012) \cite{hardcastle2012multi}&Single-ABAS&&86$^2$&&&&&&&&&&&&&&80$^2$&80$^2$&&83$^2$&&\\
La Macchia (2012) \cite{la2012systematic}&Single-ABAS&&81&69&63&&&&&&&86&&&&&78&79&81&&&\\
Zhang (2007) \cite{zhang2007automatic}&Single-ABAS&&80$^2$&&&&&&&&&85$^2$&&&&&81$^2$&80$^2$&&83$^2$&&\\
\midrule
\midrule
Radiographer (TCIA) & Manual&99.1&90.0&74.9&69.6&67.3&67.8&87.7&84.5&98.7&98.9&94.2&79.3&78.4&93.3&93.4&87.1&87.4&93.9&84.3&84.7&77.5\\
(28 scans)&&$\pm$0.2&$\pm$2.5&$\pm$10.9&$\pm$23.1&$\pm$10.4&$\pm$11.0&$\pm$8.0&$\pm$14.7&$\pm$0.7&$\pm$0.5&$\pm$2.2&$\pm$4.9&$\pm$6.2&$\pm$2.1&$\pm$1.9&$\pm$3.4&$\pm$3.1&$\pm$1.8&$\pm$4.6&$\pm$18.3&$\pm$28.5\\
\\
Our model (TCIA) & Deep Learning&98.8&85.1&80.5&81.0&64.4&63.8&81.6&75.7&98.7&98.8&92.9&77.9&76.3&92.6&93.1&84.1&84.6&91.7&80.3&81.8&77.8\\
(28 scans)&&$\pm$1.1&$\pm$7.1&$\pm$8.8&$\pm$7.2&$\pm$11.9&$\pm$9.0&$\pm$16.6&$\pm$24.5&$\pm$0.6&$\pm$0.7&$\pm$3.5&$\pm$5.0&$\pm$5.8&$\pm$2.0&$\pm$1.8&$\pm$5.8&$\pm$4.2&$\pm$1.6&$\pm$7.6&$\pm$8.7&$\pm$18.1\\
\\
\midrule
Radiographer (UCLH) & Manual&99.2&90.1&77.9&80.3&74.1&71.8&82.7&83.9&98.6&98.6&95.8&80.3&79.4&93.9&94.2&88.1&87.5&93.1&81.6&87.5&86.8\\
(21 scans)&&$\pm$0.2&$\pm$2.4&$\pm$14.0&$\pm$10.1&$\pm$7.0&$\pm$7.8&$\pm$22.6&$\pm$23.8&$\pm$0.9&$\pm$1.3&$\pm$1.2&$\pm$5.2&$\pm$7.4&$\pm$1.4&$\pm$0.9&$\pm$2.8&$\pm$3.4&$\pm$2.0&$\pm$6.0&$\pm$4.0&$\pm$4.0\\
\\
Our model (UCLH) & Deep Learning&99&91&81&79&73&72&78&81&98&98&931&77&75&95&95&85&84&93&78&83&86\\
(21 scans)&&$\pm$0.2&$\pm$2.2&$\pm$8.2&$\pm$5.7&$\pm$5.6&$\pm$5.8&$\pm$25.0&$\pm$25.8&$\pm$1.3&$\pm$2.2&$\pm$1.9&$\pm$4.8&$\pm$7.0&$\pm$1.3&$\pm$1.0&$\pm$3.8&$\pm$4.5&$\pm$1.4&$\pm$8.9&$\pm$8.4&$\pm$4.9\\
\\
Oncologist (UCLH) & Manual&99.0$^5$&91.9$^5$&68.5&75.8&63.3&61.6&86.2&87.6&98.4$^5$&98.6$^5$&95.4$^5$&77.1&76.0&94.8$^5$&94.8$^5$&90.1$^5$&90.7$^5$&94.9$^5$&87.7$^5$&91.1$^5$&90.1$^5$\\
(8 - 75 scans)$^4$&&&&$\pm$14.8&$\pm$8.5&$\pm$13.1&$\pm$14.3&$\pm$10.1&$\pm$9.9&&&&$\pm$6.3&$\pm$7.1&&&&&&&&\\
\midrule
\\
Our model (PDDCA) & Deep Learning&&84.2&&&&&&&&&93.8&71.6&69.1&&&88.1&86.6&&&76.5&79.2\\
(15 scans)&&&$\pm$5.2&&&&&&&&&$\pm$1.9&$\pm$6.2&$\pm$5.9&&&$\pm$2.0&$\pm$3.5&&&$\pm$9.1&$\pm$6.5\\
\bottomrule
\end{tabular}